\documentclass[10pt,twocolumn,letterpaper]{article}

\usepackage[pagenumbers]{cvpr} 

\usepackage{lipsum}
\usepackage{enumitem}
\usepackage{tcolorbox}
\usepackage{array}
\usepackage{multirow}
\usepackage{pifont}
\usepackage{arydshln}
\usepackage{colortbl}

\definecolor{my_green}{RGB}{51,102,0}
\definecolor{my_red}{RGB}{204, 0, 0}
\newcommand{\cmark}{\textcolor{my_green}{\ding{51}}} 
\newcommand{\xmark}{\textcolor{my_red}{\ding{55}}} 

\definecolor{customblue}{RGB}{135,206,250}

\tcbuselibrary{listings,breakable}

\newcolumntype{x}[1]{>{\centering\arraybackslash}p{#1pt}}

\newlength\savewidth

\definecolor{customblue}{RGB}{135,206,250}

\definecolor{cvprblue}{rgb}{0.21,0.49,0.74}
\usepackage[pagebackref,breaklinks,colorlinks,allcolors=cvprblue]{hyperref}

\def\paperID{xxx} 
\def\confName{CVPR}
\def\confYear{2025}

\title{PhysGame: Uncovering Physical Commonsense Violations in Gameplay Videos}

\author{Meng Cao\textsuperscript{1},~~Haoran Tang\textsuperscript{2}\footnotemark[1],~~Haoze Zhao\textsuperscript{1}\footnotemark[1],~~Hangyu Guo\textsuperscript{3},~~Jiaheng Liu\textsuperscript{3},~~Ge Zhang\textsuperscript{4},\\Ruyang Liu\textsuperscript{2},~~Qiang Sun\textsuperscript{1,5}$^{\dagger}$, ~~Ian Reid\textsuperscript{1}, ~~Xiaodan Liang\textsuperscript{1,6} \\
	\textsuperscript{1}Mohamed bin Zayed University of Artificial Intelligence~~\textsuperscript{2}Peking University\\\textsuperscript{3}Alibaba Group~~\textsuperscript{4}University of Waterloo~~\textsuperscript{5}University of Toronto~~\textsuperscript{6}Sun Yat-sen University\\
	{\small{\textsuperscript{*}Authors contributed equally to this research.~~\textsuperscript{\dag}Corresponding author.}}\\
	\textbf{\url{https://github.com/PhysGame/PhysGame}}\\
}

\begin{document}
\twocolumn[{
	\renewcommand\twocolumn[1][]{#1}
	\maketitle
	\begin{center}
	   \vspace{-0.8cm}
		\includegraphics[width=0.99\linewidth]{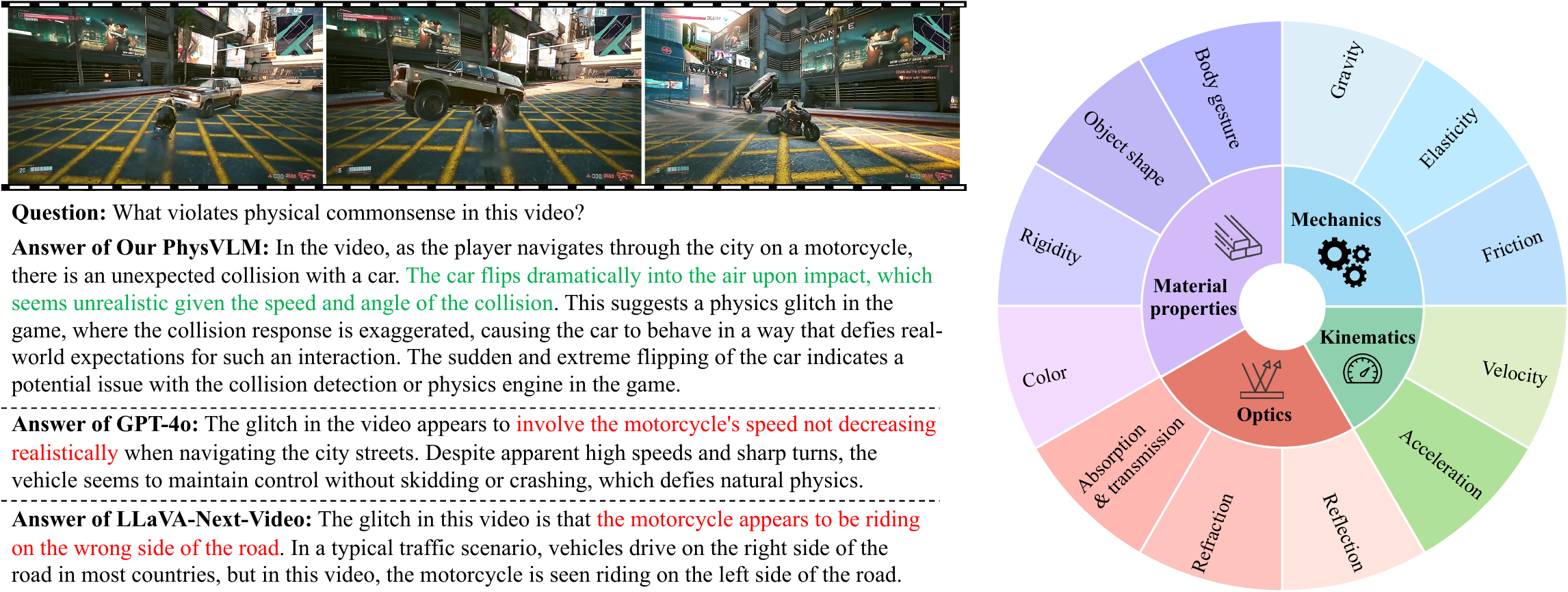}
		\captionsetup{type=figure}
		\vspace{-0.2cm}
	   \caption{\textbf{Left: Comparisons of physical commonsense understanding capability.} Our PhysVLM identifies that a motorcycle colliding and flipping a car is implausible while GPT-4o \cite{gpt4o} and LLaVA-Next-Video \cite{liu2024llavanext} fail to accurately interpret the physical commonsense violations in the video; \textbf{Right: The taxonomy of PhysGame benchmark} including 4 primary categories and 12 fine-grained sub-categories.}
		\label{fig:teaserNew}
	\end{center}
}]

\begin{abstract}
Recent advancements in video-based large language models (Video LLMs) have witnessed the emergence of diverse capabilities to reason and interpret dynamic visual content. Among them, gameplay videos stand out as a distinctive data source, often containing \emph{glitches} that defy physics commonsense. This characteristic renders them an effective benchmark for assessing the under-explored capability of physical commonsense understanding in video LLMs. In this paper, we propose \textbf{PhysGame} as a pioneering benchmark to evaluate physical commonsense violations in gameplay videos. PhysGame comprises 880 videos associated with glitches spanning four fundamental domains (\ie, mechanics, kinematics, optics, and material properties) and across 12 distinct physical commonsense. Through extensively evaluating various state-of-the-art video LLMs, our findings reveal that the performance of current open-source video LLMs significantly lags behind that of proprietary counterparts. To bridge this gap, we curate an instruction tuning dataset \textbf{PhysInstruct} with 140,057 question-answering pairs to facilitate physical commonsense learning. In addition, we also propose a preference optimization dataset \textbf{PhysDPO} with 34,358 training pairs, where the dis-preferred responses are generated conditioned on misleading titles (\ie, meta information hacking), fewer frames (\ie, temporal hacking) and lower spatial resolutions (\ie, spatial hacking). Based on the suite of datasets, we propose \textbf{PhysVLM} as a physical knowledge-enhanced video LLM. Extensive experiments on both physical-oriented benchmark PhysGame and general video understanding benchmarks demonstrate the state-of-the-art performance of PhysVLM.
\end{abstract}    
\section{Introduction} \label{sec:intro}

Large Language Models (LLMs) \cite{achiam2023gpt,reid2024gemini,brown2020language,touvron2023llama,touvron2023llama2} have achieved considerable success in understanding user instructions and delivering contextually relevant responses. Building on this foundation, video LLMs \cite{lin2023video,zhang2023video,li2023videochat,maaz2023video,li2023llama,jin2024video,munasinghe2023pg,liu2024st,lin2023mm,jin2024chat,liu2023one} have emerged as a fundamental video intelligence system by incorporating LLMs with perception and reasoning capabilities of dynamic scenes. Across the range of video types, \emph{gameplay videos} \cite{chen2021glib,ling2020using,nantes2008framework,rahman2023weak,rani2023deep,taesiri2020video,taesiri2022clip,wilkins2022learning,zheng2019wuji} present a unique challenge due to their highly dynamic and interactive environments, where agents and objects unfold in complex temporal scenes. 

Unlike real-world videos, gameplay videos frequently contain inconsistencies between the intended physics and on-screen behavior due to software bugs \cite{ling2020using,taesiri2022clip}. Typically, video glitches\footnote{We employ the term ``glitch" to describe the phenomenon of physical commonsense violation in gameplay videos.} encompass a broad spectrum of physical phenomena, thus naturally serving as a benchmark for physical commonsense understanding. Humans inherently develop an intuitive comprehension of the physical world through our experiences, enabling us to easily recognize violations of physical commonsense, even in the absence of formal physics education. For instance in Figure \ref{fig:teaserNew} (left), we can predict that the car will not be launched into the air after colliding with the motorcycle. In spite of the progress, it remains unclear how adept video LLMs are at recognizing physical commonsense violations in gameplay videos.

Although various benchmarks \cite{li2024mvbench,mangalam2023egoschema,liu2024tempcompass,fu2024video,du2024towards,wang2024lvbench,wu2024longvideobench} have been introduced to evaluate the fundamental capabilities of video LLMs, the community still lacks comprehensive evaluation standards for assessing video-based physical commonsense reasoning (\cf Table \ref{tab:comparison}). To bridge this gap, we propose \textbf{PhysGame}, a pioneering benchmark to uncover \textbf{Phys}ical commonsense violations in \textbf{Game}play videos. We focus on the intuitive adherence to the physical commonsense instead of complex physical formulas requiring explicit domain knowledge. The PhysGame benchmark consists of 880 gameplay videos containing glitches, each annotated with a high-quality multiple-choice question specifically addressing the nature of the glitch. As illustrated in Figure \ref{fig:teaserNew} (right), PhysGame spans four key physical domains (\ie, mechanics, kinematics, optics, and material properties), and encompasses 12 fine-grained categories (\eg, gravity and velocity). The video lengths range from 2.63 seconds to 239.57 seconds, covering both short clips and videos requiring long-context reasoning. 

Based on the constructed PhysGame benchmark, we provide a comprehensive analysis of state-of-the-art proprietary LLMs, \eg, GPT-4o \citep{gpt4o} and Gemini-1.5-pro \cite{reid2024gemini}, and open-source video LLMs including LLaVA-Next \cite{liu2024llavanext} and Video-LLaVA \cite{lin2023video}. Our preliminary experiments on the PhysGame benchmark reveal considerable constraints in the physical commonsense understanding capabilities of existing MLLMs: 1) Existing video LLMs exhibit limited performance on the PhysGame benchmark; 2) Open-source models tend to underperform significantly compared to proprietary counterparts, possibly due to the absence of suitable instruction-tuning dataset for physical commonsense reasoning. 

To this end, we additionally introduce the \textbf{PhysInstruct} dataset to facilitate \textbf{Phys}ical understanding oriented \textbf{Instruct}ion tuning \cite{ouyang2022training}. We curate PhysInstruct by prompting GPT-4o \cite{gpt4o} with gameplay videos and video-wise \emph{meta information} (\ie, video titles), which typically summarize key content and provide valuable hints. According to statistics, PhysInstruct consists of 140,057 instruction-following pairs regarding video glitch content. Besides, we construct the \textbf{PhysDPO} dataset for \textbf{Phys}ical \textbf{D}irect \textbf{P}reference \textbf{O}ptimization \cite{rafailov2024direct,zhang2024direct}. The \emph{preferred} responses are taken from PhysInstruct while the \emph{dis-preferred} responses are generated by prompting GPT-4o with misleading titles (\ie, meta information hacking), fewer frames (\ie, temporal hacking) and lower resolutions (\ie, spatial hacking).

Under the two-stage successive training with PhysInstruct and PhysDPO datasets, we train a \textbf{Phys}ical knowledge enhanced large \textbf{V}ideo \textbf{L}anguage \textbf{M}odel (\textbf{PhysVLM} for short). PhysVLM achieves state-of-the-art performance on the PhysGame benchmark, demonstrating its advanced capability in physical commonsense understanding. Notably, it also demonstrates impressive results on general-purpose video understanding benchmarks \cite{fu2024video,maaz2023video}. For example, PhysVLM attains an overall accuracy of 61.1\% on Video-MME \cite{fu2024video} with the use of subtitles and an average score of 3.83 on VCG benchmark \cite{maaz2023video}.

We emphasize that our focus of PhysVLM is not on introducing novel architectural designs or training strategies. Instead, we aim to integrate the existing architecture and datasets to establish a \emph{solid baseline with strong performance} on both physical commonsense and general video understanding datasets (\cf Sec.\ref{sec:4_2} and Sec.\ref{sec:4_3}). We envision that PhysVLM designed in the \emph{simple yet effective} principle will offer insights for advancing future research efforts in video LLMs.

In summary, our contributions are in three-folds:
\begin{itemize}[topsep=0pt, partopsep=0pt, leftmargin=13pt, parsep=0pt, itemsep=3pt]
    \item \emph{Physical commonsense benchmark}. The PhysGame benchmark is collected to uncover physical commonsense violations in gameplay videos.
    
    \item \emph{Instruction-following and preference-tuning datasets}. We introduce the PhysInstruct dataset to facilitate the supervised fine-tuning and the PhysDPO dataset via the meta-information, temporal, and spatial hacking for direct preference optimization.

    \item \emph{Physical knowledge enhanced Video LLM}. We propose PhysVLM, which not only demonstrates state-of-the-art performance on PhysGame but also exhibits leading performance on general video understanding benchmarks.    
\end{itemize}
\section{Related Work}\label{relatedwork}

\begin{table*}[t!]
\centering
\caption{\textbf{Comparison with existing benchmarks for video LLMs} in terms of the video number (\textbf{\#Videos}), the average video duration (\textbf{Len.}), the number of QA pair (\textbf{\#QA Pairs}), the average QA pair tokens (\textbf{QA Tokens}), the manually/automatic annotation manner (\textbf{M/A}), whether the benchmarks are gameplay video based (\textbf{Game-Bsd}), whether the questions are physical commonsense classified (\textbf{Phys-Clsf}), and whether the benchmarks contain meta information (\textbf{Meta-info}).}
\vspace{-2mm}
\scalebox{0.9}{
\begin{tabular}{lcccccccc}
\toprule
\textbf{Benchmarks} & \textbf{\#Videos}  &\textbf{Len.(s)} & \textbf{\#QA Pairs} & \textbf{QA Tokens} &  \textbf{Anno.}  & \textbf{Game-Bsd} & \textbf{Phys-Clsf}  & \textbf{Meta-info} \\
\midrule
MSRVTT-QA~\cite{xu2017video} & 2,990  & 15.2 & 72,821  & 8.4 & A & \xmark & \xmark & \xmark \\
MSVD-QA~\cite{xu2017video}& 504 & 9.8 & 13,157  & 7.6 & A & \xmark & \xmark & \xmark  \\
TGIF-QA~\cite{jang2017tgif} & 9,575 & 3.0 & 8,506  & 20.5 & A\&M & \xmark & \xmark & \xmark \\
ActivityNet-QA \cite{yu2019activitynet} & 800  & 111.4 & 8,000  & 10.2 & M & \xmark & \xmark & \xmark \\
TVQA \cite{lei2018tvqa} & 2,179 & 11.2 & 15,253  & 27.8 & M & \xmark & \xmark & \cmark\\
How2QA \cite{li2020hero}& 1,166  & 15.3 & 2,852  & 16.9 & M  & \xmark & \xmark & \cmark \\
STAR \cite{wu2star} & 914  & 11.9 & 7,098  & 19.5 & A & \xmark & \xmark  & \xmark\\
NExT-QA \cite{xiao2021next} & 1,000 & 39.5 & 8,564  & 25.3 & A & \xmark & \xmark & \xmark \\
\midrule
MVBench \cite{li2024mvbench} & 3,641  & 16.0 & 4,000  & 27.3 & A & \xmark & \xmark  & \xmark\\
Video-Bench \cite{ning2023video} & 5,917  & 56.0 & 17,036 & 21.3 & A\&M & \xmark & \xmark & \xmark\\
EgoSchema \cite{mangalam2023egoschema} & 5,063 & 180.0 & 5,063 & 126.8  & A\&M &  \xmark & \xmark & \xmark \\
AutoEval-Video \cite{chen2023autoeval} & 327 & 14.6 & 327 & 11.9 & M & \xmark & \xmark & \xmark\\
TempCompass \cite{liu2024tempcompass} & 410 & 11.4 & 7,540 & 49.2 & A\&M  & \xmark & \xmark & \xmark \\
Video-MME \cite{fu2024video} & 900 & 1017.9 & 2,700  & 35.7 & M  &  \xmark & \xmark & \cmark  \\
LVBench \cite{wang2024lvbench} &  103 & 4,101 & 1,549 & 32.0 & M & \xmark & \xmark & \xmark\\
LongVideoBench \cite{wu2024longvideobench} & 3,763 & 473.0 & 6,678 & 84.1 & A\&M & \xmark & \xmark & \xmark\\
\textbf{PhysGame (Ours)} & 880 & 25.9 & 880 & 66.9 & M & \cmark & \cmark & \cmark  \\
\bottomrule
\end{tabular}
}
\label{tab:comparison}
\end{table*}

\begin{table}[t]
\centering
\caption{\textbf{Comparison with existing gameplay video benchmarks} in terms of whether they are video-based (\textbf{Vid-Bsd}), whether they follow an instructional format (\textbf{Instruct}), and support multi-modal evaluations (\textbf{MModal}).} 
\vspace{-2mm}
\scalebox{0.9}{
\begin{tabular}{lccccc}
\toprule
\textbf{Benchmarks} & \textbf{Vid-Bsd} & \textbf{Instruct}  & \textbf{MModal} \\
\midrule
GameBunny \cite{taesiri2024videogamebunny} & \xmark & \cmark & \cmark \\
Taesiri \emph{et.al} \cite{taesiri2022clip}  & \cmark  &  \xmark & \cmark \\
GameBugDescript \cite{taesiri2022large}  & \cmark  & \cmark & \xmark \\
GlitchBench \cite{taesiri2024glitchbench}  & \xmark & \cmark & \cmark \\
\textbf{PhysGame (Ours)}  & \cmark & \cmark & \cmark \\
\bottomrule
\end{tabular}
}
\label{tab:gameComparison}
\end{table}


\noindent \textbf{Benchmarks for Video LMMs.} Integrating visual, temporal, and linguistic inputs, Video LMMs opening doors to a wide range of applications including video understanding \cite{zhang2021cola,cao2021pursuit,cao2022locvtp}, editing \cite{cao2021unifacegan,dong2024continually}, healthcare \cite{liu2023qilin,ye2023qilin}, \etc. Primarily, video LLMs have been evaluated on classical video question-answering (QA) benchmarks \cite{xu2017video,jang2017tgif,yu2019activitynet,lei2018tvqa,li2020hero,wu2star}, \eg, MSVD-QA \cite{xu2017video}, MSRVTT-QA \cite{xu2017video}, and ActivityNet-QA \cite{yu2019activitynet}. Since these benchmarks can often be addressed using a sparse set of frames \cite{wu2024longvideobench}, recent research \cite{li2024mvbench,patraucean2024perception,maaz2023video,xiao2021next,li2023vitatecs,liu2024bench} focuses on the assessment of temporal dynamics in videos. Video-ChatGPT \cite{maaz2023video} introduces a video-based generative performance benchmark, which augments videos from ActivityNet-QA \cite{yu2019activitynet} with dense descriptive captions and human annotated question-answer pairs. MVBench \cite{li2024mvbench} emphasizes temporally sensitive videos and encompasses a wide range of temporal tasks by automatically converting public annotations into multiple-choice QA formats. Recently, there has been a remarkable research interest in advancing long-form video understanding. EgoSchema \cite{mangalam2023egoschema} targets 3-minute-long egocentric videos, while MovieChat1K \cite{song2024moviechat} specializes in 10-minute-long movie videos. Recent pre-prints \cite{wang2024lvbench,fu2024video,wu2024longvideobench,du2024towards}, including Video-MME \cite{fu2024video}, LVBench \cite{wang2024lvbench}, and LongVideoBench \cite{wu2024longvideobench}, have extended the scope of long video understanding to consider more general themes and intricate reasoning tasks. Despite of the progress, few of existing benchmarks (\cf Table \ref{tab:comparison}) evaluate the physical commonsense reasoning capability in video LLMs, which acts as a critical step towards human-like video comprehension. In this paper, we bridge this gap by introducing a suite of datasets: PhysInstruct for supervised fine-tuning, PhysDPO for preference alignment, and PhysGame for evaluation.

\noindent \textbf{Gameplay Video Understanding.} Digital games \cite{hu2024survey,xu2024survey} are considered pivotal in pursuing artificial general intelligence, as they act as controllable real-world simulators and create complex problem-solving contexts. Therefore, gameplay videos are typically employed as benchmarks for evaluating the capabilities of vision-language models from the perspectives of environment perception \cite{hong2023metagpt,akoury2023framework}, context reasoning \cite{liu2023llm,wang2023avalon}, decision-making \cite{chen2023agentverse,qian2023communicative}, \etc. The majority of existing games can be classified into two categories: 1) \emph{Competition games} \cite{ma2023large,shao2024swarmbrain,hu2024pok,feng2024chessgpt,toshniwal2022chess,liemergent,gupta2023chatgpt,huang2024pokergpt,guo2023suspicion,zhang2024agent} in which players compete against one another, with the objective of outperforming others to achieve victory. Notable examples include StarCraft II \cite{ma2023large,shao2024swarmbrain}, Pokémon Battles \cite{hu2024pok}, Chess \cite{feng2024chessgpt,toshniwal2022chess,liemergent} and Poker \cite{gupta2023chatgpt,huang2024pokergpt,guo2023suspicion,zhang2024agent}. Ma et al. \cite{ma2023large} introduce TextStarCraft II, a natural language-based interface that equips LLMs with the functionality to play StarCraft II, fostering more effective reasoning and decision-making capabilities; 2) \emph{Cooperation games} \cite{carroll2019utility,gong2023mindagent,wu2021too,puigwatch,chen2024s,gan2021threedworld,puig2018virtualhome,gan1threedworld} are structured around collaboration, requiring players to work together to achieve shared goals. These games emphasize teamwork, communication, and joint problem-solving, where players must coordinate efforts to succeed and reach mutual accomplishments. In Overcooked-AI \cite{carroll2019utility}, the preparation of an onion soup requires two agents to collaborate by loading three onions into a cooking pot, thereby initiating a cooking process that spans 20-time steps.

The most relevant line of research to ours is on the topic of game bug detection \citep{taesiri2022clip,taesiri2024glitchbench,taesiri2022large,taesiri2024videogamebunny,taesiri2024searching}. Existing methods, however, either focus the classification/retrieval tasks \cite{taesiri2022clip} or limited in the static image domain \cite{taesiri2024videogamebunny,taesiri2024glitchbench} (\cf Table \ref{tab:gameComparison}). The prior work \cite{taesiri2022large} employs LLMs to detect bugs in game videos with the reliance on pre-extracted event-wise \emph{textual} descriptions and lacks support for multi-modal evaluations. In contrast, our PhysVLM addresses all these limitations and supports multi-modal instruction evaluations in videos.

\noindent \textbf{Physical Commonsense Understanding.} Even before language acquisition, children start to grasp fundamental physical commonsense by observing the properties of the world around them \cite{hespos2004conceptual}. However, acquiring such physical commonsense knowledge remains a major challenge for artificial intelligence systems. The topic of physical commonsense understanding has seen significant attention across a range of fields, \eg, visual physical reasoning \cite{lerer2016learning}, video generation \cite{meng2024towards,bansal2024videophy}, and robotics \cite{agrawal2016learning,byravan2017se3}. A collection of works \cite{lerer2016learning,battaglia2013simulation,mottaghi2016happens,chang2016compositional,fragkiadaki2015learning,battaglia2016interaction,agrawal2016learning,finn2016unsupervised,ye2018interpretable,ates2020craft} has centered on learning physical and causal reasoning through \emph{synthetic} dynamic scenarios, either from video frames \cite{finn2016unsupervised,ebert2017self,watters2017visual,lerer2016learning,mottaghi2016happens,fragkiadaki2015learning,baradelcophy,girdharcater,yiclevrer,tung2024physion++,bear1physion,chen2024compositional}, or from the symbolic environment representations \cite{battaglia2016interaction,chang2016compositional}. Within the field of video generation, recent efforts \cite{meng2024towards,bansal2024videophy} are increasingly exploring whether generative models demonstrate an understanding of intuitive physical commonsense. In robotics, learning from intuitive physics has been demonstrated effective in visuomotor planning \cite{byravan2017se3}, tool usage \cite{toussaint2018differentiable}, and construction \cite{nair2019tool}. This paper investigates video LLMs' physical commonsense reasoning skills and verifies their significance for enhancing general video understanding.
\section{Dataset \& Method}

This section details the design of the evaluation benchmark PhysGame (Sec. \ref{sec:3_1}), supervised fine-tuning (SFT) dataset PhysInstruct (Sec. \ref{sec:3_2}), and the direct preference optimization (DPO) dataset PhysDPO (Sec. \ref{sec:3_3}). The optimization procedure for PhysVLM is discussed in Section \ref{sec:3_4}.


\subsection{PhysGame Benchmark} \label{sec:3_1}

\begin{figure}[t]
	\centering
        \includegraphics[width=0.48\textwidth]{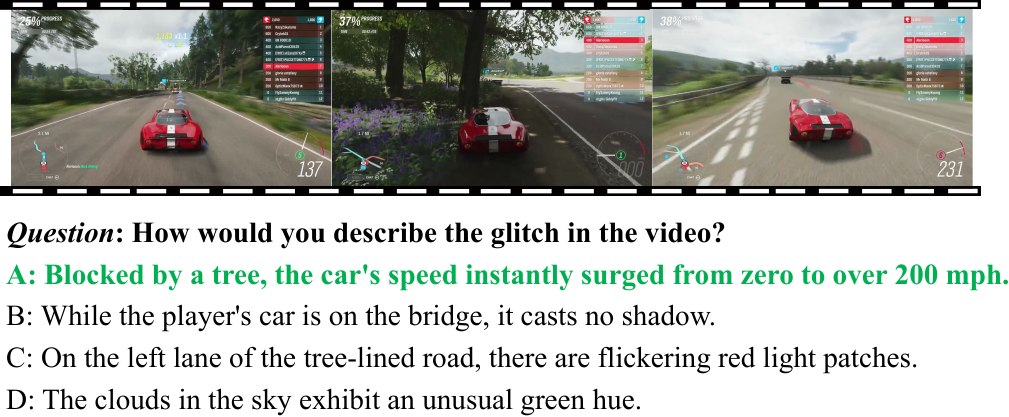}
       \vspace{-6mm}
	\caption{\textbf{The annotated multi-choice question} in PhysGame. The correct option is annotated in green.}
	\label{fig:multiChoice}
\end{figure}

Building on the intuitive cognition of physical commonsense, PhysGame benchmark introduces a comprehensive taxonomy for task categorization including four primary domains, \ie, mechanics, kinematics, optics, and material properties, and 12 fine-grained categories (\cf Figure \ref{fig:teaserNew}).

\begin{itemize}
    \item \emph{Mechanics}: This category deals with forces and torques as well as their effects on motion, which provides the foundational principles to interpret and analyze the motion of objects in videos. Typical cases include gravity, elasticity, and friction. 
    
    \item \emph{Kinematics}: This domain studies motion without considering forces, which involves fine-grained categories including velocity and acceleration over time. 
    
    \item  \emph{Optics}: It focuses on the behavior and properties of light as well as its interactions with matter. It includes reflection, refraction, and absorption \& transmission.

    \item  \emph{Material properties}: It refers to the inherent material characteristics including color, rigidity, object shape, and human body gesture. 
\end{itemize}
\begin{table}[t]
\centering
\renewcommand\arraystretch{1.1}
\caption{\textbf{The average tokens of four options} in the annotations of PhysGame benchmark.}
\vspace{-2mm}
\scalebox{0.9}{
\begin{tabular}{cx{32}x{32}x{32}x{32}}
	\toprule
        & \textbf{Opt. A} & \textbf{Opt. B} & \textbf{Opt. C} & \textbf{Opt. D}   \\
	\midrule
        Avg. tokens  & 14.40 & 14.49 & 14.46 & 14.47  \\
        \bottomrule
 \end{tabular}}
\label{tab:optionStatistic}
\end{table}
\begin{figure*}[t]
	\centering
        \includegraphics[width=0.85\textwidth]{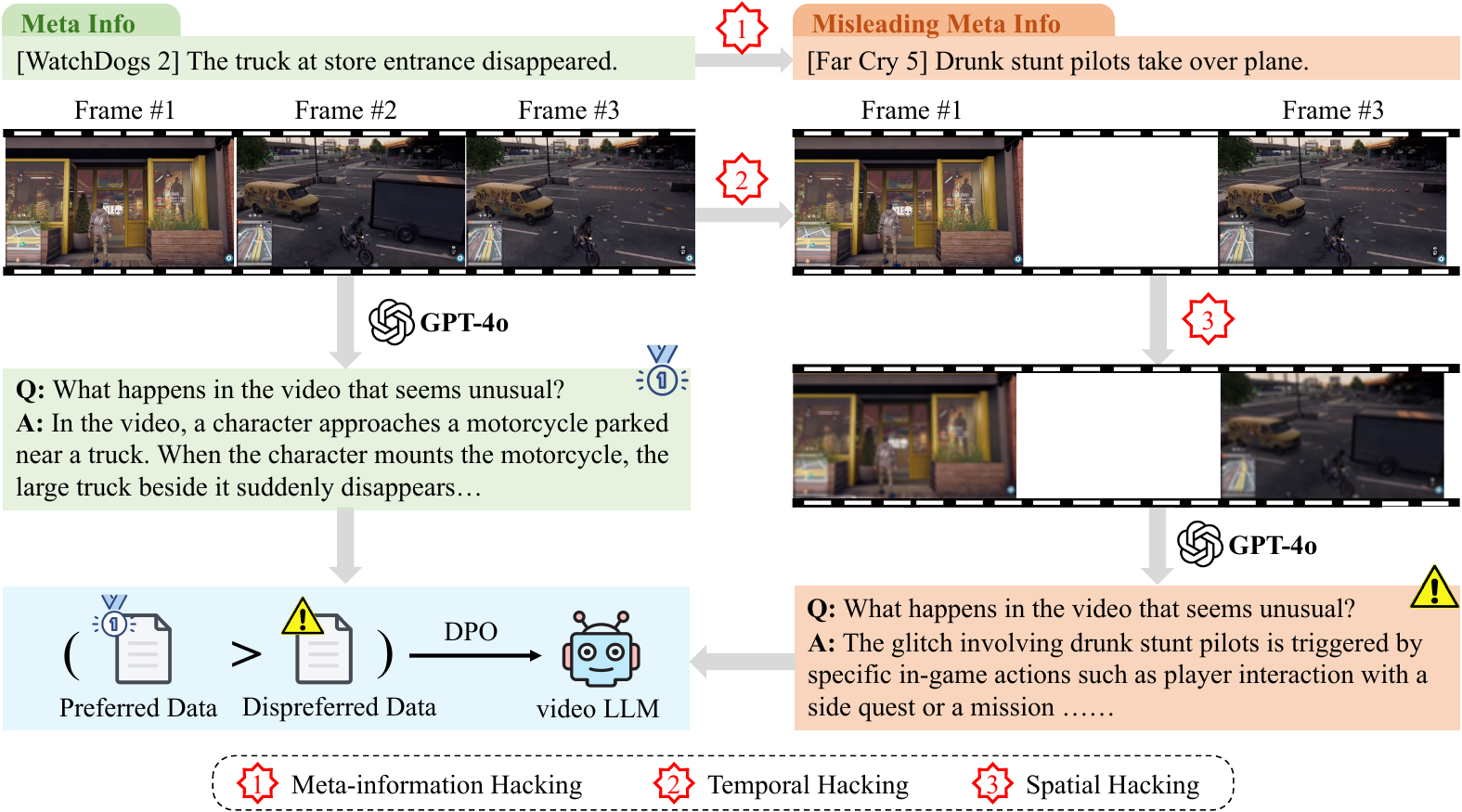}
        \vspace{-1mm}
	\caption{\textbf{Overview of the direct preference optimization training}, where the preferred data is generated with the guidance of associated meta-information (\ie, title) while dispreferred data is generated with misleading titles (\ie, meta-information hacking), fewer frames (\ie, temporal hacking) and lower resolutions (\ie, spatial hacking).}
	\label{fig:pipeline}
\end{figure*}

\noindent \textbf{Video Collection and Filter.} The videos in PhysGame are mainly crawled from the reddit page\footnote{\url{www.reddit.com/r/GamePhysics/}} which contains gameplay videos with unusual events and glitches. To balance different categories, we also augment videos from YouTube via keyword searching. We conduct manual checks based on the following two criteria: 1) \emph{Duplicate check}: The Reddit discussion forum may feature multiple references to the same video, resulting in duplicate downloading. We manually check to confirm that each video in PhysGame is distinct; 2) \emph{Content check}: The pool of downloaded videos may incorporate non-game elements, which we rigorously filter out of our PhysGame benchmark.

\noindent \textbf{Annotation Scheme.} Based on the collected gameplay videos, we create the question-answer pairs in a four-way multiple-choice format to facilitate convenient evaluation. Specifically, the correct options describe the video-specific glitches that contravene physical commonsense principles. It is important to note that some videos may exhibit multiple glitches. We therefore instruct expert annotators to review the entire video to ensure all the appearing glitches are included in the correct answer. 

To enhance the plausibility of the distractor options, we have provided expert annotators with three guiding principles: \textbf{1)} Instead of imagining arbitrary glitches, the glitch in the distractor options should be highly correlated to the individuals or actions observed in the videos. For example in Figure \ref{fig:multiChoice}, the distractor option B includes \texttt{car} and \texttt{shadow} that are genuinely present in the video. This annotation principle forces video LLMs to comprehend the glitchy content, rather than merely selecting answers by identifying contained objects or actions; \textbf{2)} The four choice options should be of similar length, which helps prevent any preference biases in video LLMs. From Table \ref{tab:optionStatistic}, it can be observed that all four options exhibit comparable token numbers; \textbf{3)} To mitigate choice bias, the distribution of the correct option among the four choices should be equitable, (\ie, 25\% likelihood for each option). 

\noindent \textbf{Quality Control.} To guarantee the quality of our dataset, we conduct a two-fold quality control process including human inspection and automatic LLM-assisted inspection: \textbf{1)} All the initially annotated question-answering pairs undergo rigorous cross-inspection by different human annotators. For the correct options, the inspectors must assess whether they comprehensively and accurately describe all instances of physical commonsense violations present. For the distractor options, the inspectors are required to evaluate whether they are sufficiently deceptive, specifically by including objects or actions depicted in the video; \textbf{2)} We exclude question-answering pairs that can be correctly answered by GPT-4o \cite{gpt4o} solely based on the question and options without the need to view the video. By statistics, we limit the accuracy of GPT-4o in the question-only scenario to less than 25\%. Through the rigorous construction and review process, we present the PhysGame benchmark which is of high quality and well-balanced. The specific statistics are available in Table \ref{tab:comparison}.

\subsection{PhysInstruct Dataset} \label{sec:3_2}

To improve the physical commonsense understanding ability of video LLMs, we develop the PhysInstruct dataset for supervised fine-tuning. The video collection procedure follows the same process as that in PhysGame. To prevent data leakage, we diligently exclude any videos included in PhysGame. We follow the self-instruction paradigm \cite{li2024llava} to construct PhysInstruct by prompting GPT-4o \cite{gpt4o}. 

In terms of the instruction generation\footnote{Refer to the supplementary material for detailed prompts.\label{prompt}}, we aspire for the questions to be as diverse as possible. We set up three question types, varying from direct to indirect assessment of the glitches in videos: i) explicitly inquiring about glitches in the video, ii) probing anomalies present in the video, iii) or merely straightforward questions regarding the video content. We adopt in-context learning \cite{dong2022survey} with three examples corresponding to the mentioned three scenarios: i) \emph{What is the description of the glitch observed in the video?} ii) \emph{Are there any abnormalities present in the videos?} iii) \emph{Please provide a description of the video content.} 

\begin{figure}[t]
	\centering
    \includegraphics[width=0.49\textwidth]{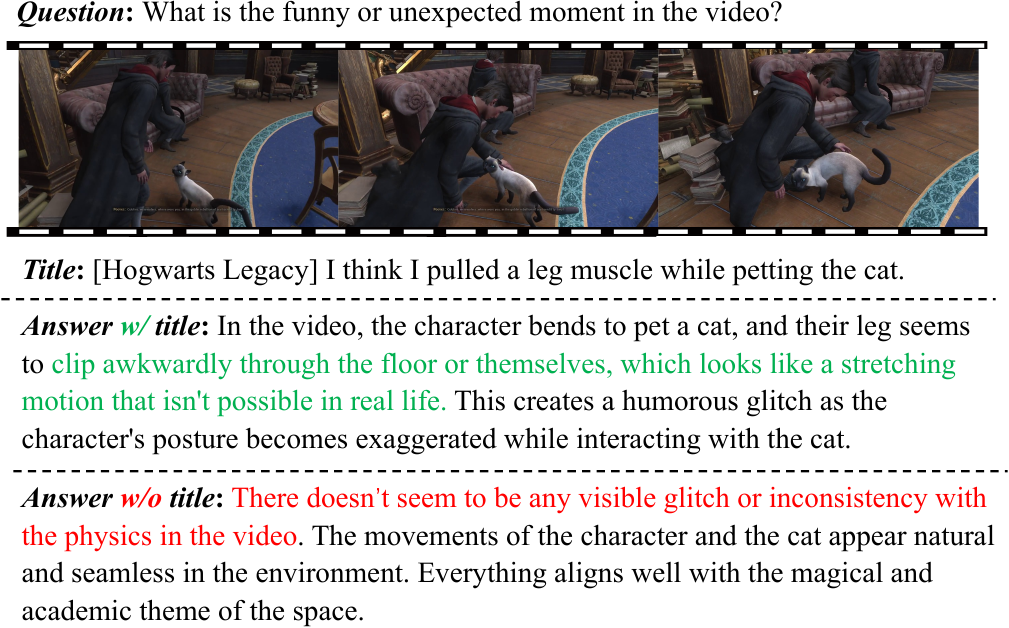}
        \vspace{-5mm}
	\caption{\textbf{Example cases in the PhysInstruct dataset} with (\emph{w/}) or without (\emph{w/o}) meta-information hints.}
	\label{fig:compareTitle}
\end{figure}

As for the response generation\footref{prompt}, the preliminary experiments suggest that the intuitive prompting method leads to remarkable errors, regardless of how we adjust the prompt contents. To alleviate this, we propose a \emph{meta-information guided prompting} strategy. Specifically, we found that the meta-information (\eg, title) associated with each video offers insight into the fundamental content. For example in Figure \ref{fig:compareTitle}, the title indicates the glitch concerning \texttt{leg} \texttt{muscle}. Therefore, we propose to incorporate video-wise meta-information in the prompt, resulting in more accurate instruction-following generation (\cf Figure \ref{fig:compareTitle}). In Figure \ref{fig:compareTitle}, the meta-information helps GPT-4o to detect the \texttt{leg} gesture glitch, while its absence leads to the degradation of physical commonsense understanding. In total, we generate 140,057 instruction-following pairs.

\subsection{PhysDPO Dataset} \label{sec:3_3}

We construct the preference alignment dataset PhysDPO to deliver more trustworthy and reliable responses. As shown in Figure \ref{fig:pipeline}, we regard the generated answers in the PhysInstruct dataset as the preferred responses and the dis-preferred responses are generated via the combination of meta-information hacking, temporal hacking, and spatial hacking. We prompt GPT-4o with the misleading meta-information as well as video frames with the reduced frame count and decreased frame resolution\footref{prompt}. Specifically, the dis-preferred data are generated based on one single frame with both height and width reduced to 1/16 of the original dimensions. Through the above generation process, we compile the PhysDPO dataset with 34,358 training pairs.

\subsection{PhysVLM} \label{sec:3_4}

Through our experiments in Sec.\ref{sec:4_2}, we observe that the capabilities of current open-source models are markedly inferior to those of proprietary models. To establish a strong open-source baseline for physical commonsense understanding, we propose PhysVLM by employing cutting-edge architecture and high-quality training datasets. 

\noindent \textbf{Architecture.} For the architecture design, we primarily adopt PPLLaVA \cite{liu2024ppllava}, solely substituting the Vicuna-7B \cite{vicuna2023} LLM with the high-performing Qwen2-7B \cite{yang2024qwen2}. 

\noindent \textbf{Supervised Fine-tuning.} We introduce the hybrid training datasets that include our physics-oriented datasets alongside general image and video datasets. Following LLaVA-NeXT Interleave \cite{li2024llavainterleave}, we utilize a combined dataset comprising 300K images randomly sampled from the LLaVA-1.5 \cite{liu2024visual} training dataset, 300K video samples in LLava-Hound \cite{zhang2024direct}, and our PhysInstruct dataset, totaling around 740k training samples. We apply the conventional auto-regressive loss in this stage.

\noindent \textbf{Direct Preference Optimization.} We use the preferred and dis-preferred data in both LLaVA-Hound-DPO-17k \cite{zhang2024direct} and our self-constructed PhysDPO in this stage, leading to a total of 51k training samples. The training loss follows the standard methodology \cite{zhang2024direct,rafailov2024direct}.
\begin{table*}
\centering
\caption{\textbf{Evaluation results (\%) of open-source and proprietary multi-modal LLMs on PhysGame.} The fine-grained categories include gravity, elasticity, friction, velocity, acceleration, reflection, refraction, absorption \& transmission, color, rigidity, object shape, and body gesture. AVG denotes the average accuracy. PhysVLM-SFT denotes PhysVLM only undergoes supervised fine-tuning while PhysVLM-DPO denotes PhysVLM with consecutive supervised fine-tuning and direct preference optimization.}
\vspace{-2mm}
\label{tab:resPhys}
\scalebox{0.9}{
\begin{tabular}{llx{22}x{20}x{20}x{20}x{20}x{20}x{20}x{20}x{20}x{20}x{20}x{20}x{20}}
\toprule
\multicolumn{1}{c}{\multirow{2}{*}{\textbf{Models}}} & & \multicolumn{1}{c|}{\multirow{2}{*}{\textbf{AVG}}} & \multicolumn{3}{c|}{\textbf{Mechanics}} & \multicolumn{2}{c|}{\textbf{Kinematics}} & \multicolumn{3}{c}{\textbf{Optics}} & \multicolumn{4}{c}{\textbf{Material}}  \\ 
\cmidrule(r){4-6} \cmidrule(r){7-8} \cmidrule(r){9-11} \cmidrule(r){12-15}
& & & {Grav.}    & Elast.  & Fric.  & Velo.    & Acc. & Refl.   & Refr. & Abs. & Col. & Rig. & Sha. & Gest.  \\ 
\midrule
\multicolumn{15}{c}{\emph{\textcolor{blue}{\textbf{Proprietary Multi-modal LLMs}}}}  \\ 
Claude3.5-Sonnet & \cite{anthropic2024claude} & 54.3 & \textbf{50.7} & 58.8 & 50.6 & \textbf{53.2} & 59.1 & \textbf{50.0} & 50.0 & 49.2 & 64.4 & 52.7 & 50.0 & \textbf{62.1} \\ 
Claude3.5-SonnetV2& \cite{anthropic2024claude} & 47.6 & 46.5 & 52.5 & 46.6 & 37.2 & 53.4 & 47.8 & 50.0 & 33.9 & 55.6 & 54.1 & 43.8 & 51.7\\
Gemini-1.5-pro &\cite{team2024gemini} & 55.2 & \textbf{50.7} & \textbf{70.0} & 48.9 & 51.1 & 59.1 & \textbf{50.0} & 42.9 & \textbf{52.5} & \textbf{71.1} & \textbf{56.8} & 53.1 & 58.6 \\
Gemini-1.5-pro-flash &\cite{team2024gemini} & 48.5 & 47.9 & 52.5 & 51.7 & 43.6 & 51.1 & 43.5 & 53.6 & 33.9 & 64.4 & 43.2 & 46.9 & 49.4 \\
GPT-4V &\cite{achiam2023gpt} & 45.9 & 40.8 & 60.0 & 48.3 & 34.0 & 48.9 & 43.5 & 46.4 & 42.4 & 53.3 & 45.9 & 37.5 & 44.8 \\
GPT-4o-0806 &\cite{gpt4o} & \textbf{56.1} & 47.9 & 61.3 & \textbf{59.1} & 43.6 & \textbf{61.4} & 43.5 & 53.6 & 50.8 & 68.9 & 54.1 & \textbf{65.6} & 63.2 \\
GPT-4o-mini-0718 &\cite{gpt4o} & 40.3 & 43.7 & 43.8 & 39.2 & 35.1 & 44.3 & 30.4 & 46.4 & 42.4 & 44.4 & 37.8 & 37.5 & 41.4 \\
Qwen-VL-max &\cite{bai2023qwen} & 50.9 & \textbf{50.7} & 53.8 & 51.1 & 31.9 & 46.6 & \textbf{50.0} & \textbf{60.7} & 50.8 & 64.4 & 48.6 & \textbf{65.6} & 59.8 \\
\midrule
\multicolumn{15}{c}{\emph{\textcolor{blue}{\textbf{Open-source Multi-modal LLMs}}}}  \\ 
LLaVA-Next-Video &\cite{liu2024llavanext} & 32.2 & 43.7 & 33.8 & 27.3 & 34.0 & 22.7 & 21.7 & 35.7 & 23.7 & 35.6 & 41.9 & 34.4 & 37.9 \\
Video-LLaVA &\cite{lin2023video} & 29.0 & 32.4 & 22.5 & 27.8 & 31.9 & 26.1 & 19.6 & 35.7 & 32.2 & 31.1 & 36.5 & 28.1 & 27.6 \\
LLaVA-OneVision &\cite{li2024llava} & 47.7 & 50.7 & 50.0 & 46.0 & 39.4 & 45.5 & 43.5 & \textbf{71.4} & \textbf{40.7} & 55.6 & 44.6 & \textbf{56.2} & 52.9 \\
InternVL2 &\cite{chen2024far} & 33.4 & 29.6 & 31.2 & 38.6 & 35.1 & 30.7 & 30.4 & 53.6 & 35.6 & 26.7 & 29.7 & 18.8 & 34.5 \\
VideoChat2 & \cite{li2024mvbench} & 34.3 & 33.8 & 35.0 & 29.5 & 41.5 & 28.4 & 28.3 & 32.1 & 33.9 & 33.3 & 41.9 & 21.9 & 44.8 \\
ST-LLM & \cite{liu2024st} & 32.8 & 32.4 & 26.2 & 26.7 & 37.2 & 28.4 & 37.0 & 25.0 & 28.8 & 33.3 & 40.5 & 37.5 & 46.0 \\
Chat-UniVi & \cite{jin2024chat} & 29.5 & 28.2 & 27.5 & 29.5 & 39.4 & 23.9 & 28.3 & 32.1 & 30.5 & 31.1 & 18.9 & 28.1 & 35.6 \\
PPLLaVA &  \cite{liu2024ppllava} & 38.4 & 45.1 & 38.8 & 42.6 & 30.9 & 30.7 & 41.3 & 39.3 & 35.6 & 44.4 & 39.2 & 18.8 & 43.7 \\
\textbf{PhysVLM-SFT} &  & 56.7 & 54.9 & 62.5 & \textbf{60.2} & 51.1 & \textbf{63.6} & \textbf{45.7} & 57.1  & 28.8 & \textbf{64.4} & 51.4 & 50.0 & 72.4 \\
\textbf{PhysVLM-DPO} &  & \textbf{59.5} & \textbf{64.8} & \textbf{66.3} & \textbf{60.2} & \textbf{59.6} & 60.2 & 39.1 & 67.9  & 35.6 & 57.8 & \textbf{62.2} & 37.5 & \textbf{78.2} \\
\bottomrule
\end{tabular}
}
\end{table*}
\begin{table*}
\centering
\caption{\textbf{Evaluation results (\%) on Video-MME.} ``\emph{w/} subs'' and ``\emph{w/o} subs'' respectively denote ``with subtitles'' and ``without subtitles''.} 
\vspace{-1mm}
\label{tab:resVideoMME}
\scalebox{0.88}{
\begin{tabular}{ll!{\vrule width \lightrulewidth}c!{\vrule width \lightrulewidth}cccccccc} 
\toprule
\multicolumn{1}{c|}{\multirow{2}{*}{\textbf{Models}}} & & \multirow{2}{*}{\begin{tabular}[c]{@{}c@{}}\textbf{LLM}\\\textbf{Params}~~\end{tabular}} & \multicolumn{2}{c}{\textbf{Short (\%)}} & \multicolumn{2}{c}{\textbf{Medium (\%)}} & \multicolumn{2}{c}{\textbf{Long (\%)}} & \multicolumn{2}{c}{\textbf{Overall (\%)}}  \\ 
& & & \emph{w/o} subs    & \emph{w/} subs     & \emph{w/o} subs    & \emph{w/} subs    & \emph{w/o} subs   & \emph{w/} subs  & \emph{w/o} subs & \emph{w/} subs \\ 
\midrule
\textcolor{gray}{InternVL-Chat-V1.5} & \cite{chen2024far} & \textcolor{gray}{20B}  & \textcolor{gray}{60.2} & \textcolor{gray}{61.7}  & \textcolor{gray}{46.4} & \textcolor{gray}{49.1} & \textcolor{gray}{45.6} & \textcolor{gray}{46.6}  & \textcolor{gray}{50.7}   & \textcolor{gray}{52.4} \\ 
\textcolor{gray}{LLaVA-NeXT-Video} & \cite{liu2024llavanext}  & \textcolor{gray}{34B}  & \textcolor{gray}{61.7} & \textcolor{gray}{65.1}   & \textcolor{gray}{50.1}  & \textcolor{gray}{52.2} & \textcolor{gray}{44.3} & \textcolor{gray}{47.2} & \textcolor{gray}{52.0} & \textcolor{gray}{54.9}   \\
\textcolor{gray}{VILA-1.5} & \cite{lin2024vila} &  \textcolor{gray}{34B} & \textcolor{gray}{68.1} & \textcolor{gray}{68.9} & \textcolor{gray}{58.1} & \textcolor{gray}{57.4} & \textcolor{gray}{50.8} & \textcolor{gray}{52.0}  & \textcolor{gray}{59.0} & \textcolor{gray}{59.4}  \\ 
\textcolor{gray}{LLaVA-OneVision} & \cite{li2024llava} & \textcolor{gray}{72B} & \textcolor{gray}{76.7} & \textcolor{gray}{79.3} & \textcolor{gray}{62.2} & \textcolor{gray}{66.9} & \textcolor{gray}{60.0} & \textcolor{gray}{62.4} & \textcolor{gray}{66.3} & \textcolor{gray}{69.6} \\
Qwen-VL-Chat  &\cite{bai2023qwen} & 7B & 46.9 & 47.3 & 38.7 & 40.4 & 37.8 & 37.9 & 41.1 & 41.9 \\
Video-LLaVA &\cite{lin2023video}   & 7B  & 45.3  & 46.1  & 38.0  & 40.7  & 36.2 & 38.1 & 39.9 & 41.6 \\
ST-LLM & \cite{liu2023one}   & 7B   & 45.7 & 48.4  & 36.8  & 41.4   & 31.3  & 36.9  & 37.9 & 42.3  \\
VideoChat2-Mistral & \cite{li2024mvbench}  & 7B & 48.3  & 52.8  & 37.0 & 39.4   & 33.2 & 39.2 & 39.5 & 43.8  \\
Chat-UniVi-V1.5 & \cite{jin2024chat}   & 7B  & 45.7  & 51.2   & 40.3   & 44.6  & 35.8  & 41.8 & 40.6 & 45.9 \\
LLaVA-NeXT-Video & \cite{liu2024llavanext} & 7B & 45.9 & 49.8 & 40.3 & 44.3 & 36.6 & 41.0 & 40.9 & 45.0 \\
PPLLaVA & \cite{liu2024ppllava} & 7B & 58.7 & 62.8 & 45.6 & 50.4 & 42.2 & 47.4 & 48.8 & 53.6 \\
\textbf{PhysVLM-SFT} & & 7B  & 64.1 & 68.0 & \textbf{55.0} &  \textbf{61.7} & 46.4 & 50.3 & 55.2 & 60.0 \\
\rowcolor{customblue}\textbf{PhysVLM-DPO} & & 7B &  \textbf{66.1} & \textbf{70.0}  & 54.3 & 59.6 & \textbf{47.1} & \textbf{53.8}  & \textbf{55.8} & \textbf{61.1} \\
\bottomrule
\end{tabular}
}
\end{table*}
\section{Experiments}

\subsection{Experimental Settings}

\noindent \textbf{Implementation Details.} For SFT training, the pooling kernel and strides are set to $(1, 3, 3)$ for image inputs and $(2, 3, 3)$ for video inputs. PhysVLM lasts for one epoch with a batch size of 256 and a learning rate of 2e-5. The input frame is set to 32 following PPLLaVA \cite{liu2024ppllava}. For DPO training, the pooling kernel and strides are set to $(1, 3, 3)$, and the input frame is set to 16. The DPO training lasts two epochs with a batch size of 64 and a learning rate of $5\mathrm{e}{-6}$. All experiments are conducted on 8 NVIDIA A100 GPUs.

\subsection{Evaluations on PhysGame} \label{sec:4_2}

\noindent \textbf{Evaluation Settings.} We benchmark PhysGame on 8 proprietary multi-modal LLMs, \ie, Claude3.5-Sonnet \cite{anthropic2024claude}, Claude3.5-SonnetV2 \cite{anthropic2024claude}, Gemini-1.5-pro \cite{team2024gemini}, Gemini-1.5-pro-flash \cite{team2024gemini}, GPT-4V \cite{achiam2023gpt}, GPT-4o-0806 \cite{gpt4o}, GPT-4o-mini-0718 \cite{gpt4o} and Qwen-VL-max \cite{bai2023qwen}, as well as 8 open-source models including LLaVA-Next-Video \cite{liu2024llavanext}, Video-LLaVA \cite{lin2023video},  LLaVA-OneVision \cite{li2024llava}, InternVL2 \cite{chen2024far}, VideoChat2 \cite{li2024mvbench}, ST-LLM \cite{liu2024st}, Chat-UniVi \cite{jin2024chat} and PPLLaVA\cite{liu2024ppllava}. We follow Video-MME \cite{fu2024video} to utilize the official frame configurations provided for each video LLM. We employ accuracy as the evaluation metric for our curated multi-choice questions. The evaluation prompt is available in the supplementary material.

\noindent \textbf{Performance Analysis.} The evaluation results on the PhysGame benchmark are demonstrated in Table \ref{tab:resPhys}. Among all proprietary models, GPT-4o and Gemini-1.5-pro demonstrate the best performance, achieving average accuracy scores of 56.1\% and 55.2\%, respectively. Across all the fine-grained domains, GPT-4o achieves superior performance in friction and acceleration. In contrast, Gemini-1.5-pro shows a stronger capability in understanding physical commonsense related to gravity, elasticity, reflection, absorption \& transmission, color, and rigidity.

Furthermore, existing open-source models fall significantly behind proprietary models. Even the best-performing open-source model, LLaVA-OneVision, reaches only 47.7\% average accuracy. In comparison, our proposed PhysVLM achieved state-of-the-art performance among all proprietary and open-source models. Compared to open-source methods, our PhysVLM attains the highest performance in 6 domains out of the total 12 evaluated domains. Notably, PhysVLM-DPO surpasses the best-performing proprietary model GPT-4o by an absolute margin of 3.4\% in the metric of average accuracy.

\begin{table}[t]
\centering
\caption{\textbf{Evaluation results on VCG benchmark \cite{maaz2023video}.} Methods marked by $^{*}$ use DPO or PPO \cite{schulman2017proximal}. CI, DO, CU, TU, and CO respectively denote correctness of information, detail orientation, contextual understanding, temporal understanding, and consistency. AVG is the average result.}
\vspace{-2mm}
\scalebox{0.86}{
\begin{tabular}{lcccccc}
	\toprule
        \textbf{Methods}   & \textbf{CI} & \textbf{DO} & \textbf{CU} & \textbf{TU}   & \textbf{CO} & \textbf{AVG} \\
	\midrule
        VideoChat  & 2.23 & 2.50 & 2.53 & 1.94 & 2.24 & 2.29 \\
        Video-ChatGPT & 2.50 & 2.57 & 2.69 & 2.16 & 2.20 & 2.42 \\
        BT-Adapter & 2.68 & 2.69 & 3.27 & 2.34 & 2.46 & 2.69 \\
        Chat-UniVi & 2.89 & 2.91 & 3.46 & 2.89 & 2.81 & 2.99 \\
        VideoChat2 & 3.02 & 2.88 & 3.51 & 2.66 & 2.81 & 2.98 \\
        LLaMA-VID  & 2.96 & 3.00 & 3.53 & 2.46 & 2.51 & 2.89 \\
        ST-LLM     & 3.23 & 3.05 & 3.74 & 2.93 & 2.81 & 3.15 \\
        PLLaVA     & 3.21 & 2.86 & 3.62 & 2.33 & 2.93 & 2.99 \\
        LLaVA-Next-Video & 3.39 & \textbf{3.29} & \textbf{3.92} & 2.60 & 3.12 & 3.26 \\
        PPLLaVA & 3.32 & 3.20 & 3.88 & \textbf{3.00} & 3.20 & 3.32 \\
        \textbf{PhysVLM-SFT} & \textbf{3.59} & 3.07	& 3.89	& 2.74 & \textbf{3.44} & \textbf{3.35} \\
        \midrule
        LLaVA-Next-Video$^{*}$ & 3.64 & 3.45 & 4.17 & 2.95 & 4.08 & 3.66 \\
        PPLLaVA$^{*}$ & 3.85 & 3.56 & 4.21 & \textbf{3.21} & 3.81 & 3.73 \\
        \textbf{PhysVLM-DPO$^{*}$} & \textbf{3.89} & \textbf{3.69} & \textbf{4.26} & 3.11 & \textbf{4.19} & \textbf{3.83} \\
        \bottomrule
\end{tabular}}
\label{tab:resVCG}
\end{table}

\subsection{Evaluations of General Video Understanding} \label{sec:4_3}

To further demonstrate the generalizability of our model, we conducted experiments on general video LLM benchmarks including Video-MME \cite{fu2024video} and VCG benchmarks \cite{maaz2023video}. We follow the common practice by using \texttt{GPT-3.5-turbo0613} version for the evaluation of the VCG benchmark. Video-MME is in the format of multi-choice questions and thus the evaluation is more objective by eliminating the reliance on GPT. 

The comparison results on the Video-MME benchmark are demonstrated in Table \ref{tab:resVideoMME}. Our PhysVLM achieves superior performance among all the 7-B models. Surprisingly, as the 7B model, both PhysVLM-SFT and PhysVLM-DPO outperform the 34B model LLaVA-NeXT-Video by 3.2\% and 3.8\% absolute improvements on the overall performance without using subtitles. By comparing PhysVLM-SFT and PhysVLM-DPO, we find that DPO training using the proposed PhysDPO data results in performance gains on both short and long videos, while performance on medium-length videos experiences a slight decline. 

We summarize the results on the VCG benchmark in Table \ref{tab:resVCG}. In the case of models using only SFT, our PhysVLM-SFT achieves the best performance in terms of the average score. In the evaluation across four subcategories, PhysVLM-SFT performs exceptionally well in the correctness of information and consistency categories. Compared to PPLLaVA and LLaVA-Next-Video which use DPO or PPO training, our PhysVLM-DPO also demonstrates superior performance, further validating the outstanding capabilities of the proposed PhysVLM model in general video understanding.

\subsection{Ablations}
\begin{table}[t]
\centering
\begin{minipage}{0.25\textwidth}
\scalebox{0.85}{
\begin{tabular}{lccccc}
	\toprule
        \textbf{Methods} & \textbf{AVG}     \\
	\midrule
        PhysVLM-DPO &  \textbf{59.5}  \\
        \emph{w/o} temporal hacking & 57.6  \\
        \emph{w/o} spatial hacking &  57.3  \\
        \emph{w/o} meta-info hacking &  57.4  \\
        \bottomrule
\end{tabular}
}
\end{minipage}
\begin{minipage}{0.20\textwidth} 
\caption{\textbf{Ablation studies of the temporal, spatial, and meta-info hacking} in the PhysDPO dataset generation process.}
	\label{tab:ablate}
	\end{minipage}
\end{table}
\begin{table}
\centering
\setlength{\tabcolsep}{1.5mm}{} 
\caption{\textbf{Ablations of training data} in SFT and DPO stages. AVG denotes the average accuracy on the PhysGame benchmark.} 
\vspace{-2mm}
\label{tab:ablateData}
\scalebox{0.85}{
\begin{tabular}{x{30}lx{35}}
\toprule
\textbf{Stage} & \textbf{Training Data} & \textbf{AVG} \\
\midrule
SFT & LLava-Hound   &  40.7 \\
SFT & LLava-Hound\cite{zhang2024direct}, LLaVA-Image \cite{liu2024visual} & 46.0\\
SFT & LLava-Hound, LLaVA-Image, \textbf{PhysInstruct} & \textbf{56.7} \\
\midrule
DPO & LLava-Hound-DPO \cite{zhang2024direct} & 52.9 \\
DPO & LLava-Hound-DPO, \textbf{PhysDPO} & \textbf{59.5} \\
\bottomrule
\end{tabular}
}
\end{table}

\noindent \textbf{Ablations of DPO Dataset Generation.} In Section \ref{sec:3_3}, the dis-preferred responses in PhysDPO are generated by prompting GPT-4o with misleading titles (\ie, meta information hacking), fewer frames (\ie, temporal hacking) and lower spatial resolutions (\ie, spatial hacking). We ablate on these three kinds of hacking in Table \ref{tab:ablate}. We found that removing any one of the three components leads to a decline in overall performance. For instance, omitting temporal hacking results in a 1.9\% decrease in the final PhysVLM performance on the AVG metric. 

\noindent \textbf{Ablations of Training Data.} We use the hybrid training dataset in both the SFT and DPO training stages. Here we ablate on each dataset to investigate the specific contributions. We report the average accuracy on the PhysGame benchmark in Table \ref{tab:ablateData}. The comparison results demonstrate that introducing PhysInstruct and PhysDPO respectively leads to 10.7\% and 6.6\% performance boosts, further validating the effectiveness of the curated datasets. More ablation results on Video-MME and VCG benchmarks are available in the supplementary material.
\section{Conclusion}
This paper investigates current video LLMs' understanding of physical commonsense in gameplay videos. To achieve this, we introduce PhysGame benchmark, consisting of glitchy gameplay videos accompanied by annotated question-answer pairs to identify and analyze physical commonsense violations. The extensive experiments reveal that the performance of open-source models falls significantly behind that of proprietary counterparts. To this end, we propose PhysVLM as an open-source physical-knowledge-enhanced video LLM. To facilitate training, we curate a suite of datasets including PhysInstruct for instruction tuning and PhysDPO for preference alignment. Experiments manifest that PhysVLM achieves state-of-the-art performance on both physical-oriented benchmark PhysGame and general video understanding benchmarks.

{
    \small
    \bibliographystyle{ieeenat_fullname}
    \bibliography{main}
}

\clearpage
\section{Supplementary Material} \label{sec:6}

This supplementary material is organized as follows. We firstly clarify our motivation in designing the proposed benchmark and methodology in Sec \ref{sec:6_1}. Then, we present more ablation studies in Sec \ref{sec:6_2}. The qualitative comparison results are illustrated in Sec \ref{sec:6_3}. Finally, the prompts for supervised fine-tuning (SFT), direct preference optimization (DPO), and evaluation are detailed in Sec \ref{sec:6_4}.

\subsection{Motivation Clarification} \label{sec:6_1}

\noindent \textbf{Q:} \emph{Why use gameplay videos rather than real-world videos for benchmarking physical commonsense understanding?}

\noindent \textbf{A:} Compared to real-world videos, gameplay videos offer several advantages for physical commonsense benchmarks: 1) \textbf{Easier to define:} Instead of complex physical formulas, this paper focuses on the intuitive adherence to the physical commonsense. Given real-world videos, it is both \emph{challenging} and \emph{unnecessary} to exhaustively cover and interpret all normal physical phenomena. In contrast, gameplay videos typically contain glitches that violate physical commonsense. This can simplify the definition and evaluation of the physical commonsense understanding, \ie, focusing on interpreting physical commonsense violation rather than trying to enumerate all the existing normal physical phenomena; 2) \textbf{More meaningful:} The video game industry generates substantial annual revenue with billions of gamers \cite{pashkov2021video}. Automatically detecting in-game glitches acts as a highly demanding task for gameplay video stress testing. Therefore, developing video LLMs to uncover physical commonsense violations in gameplay videos may potentially offer one automatic and end-to-end solution.

\noindent \textbf{Q:} \emph{Differences from prior gameplay glitch datasets.}

\noindent \textbf{A:} We have briefly clarified the differences between our proposed PhysGame benchmark and existing gameplay glitch datasets in the related work section of the main paper. Here, we provide more detailed discussions. 

The overall comparisons with existing benchmarks are illustrated in Table \ref{tab:gameComparison}. Taesiri \emph{et.al} \cite{taesiri2022clip} focus on gameplay video retrieval by leveraging the zero-shot transfer capabilities of the large-scale pre-trained CLIP model \cite{radford2021learning}. Several existing works are limited in the static image domain \cite{taesiri2024videogamebunny,taesiri2024glitchbench}. GlitchBench \cite{taesiri2024glitchbench} is proposed for video-game quality assurance by evaluating the reasoning capabilities of LMMs under unusual and glitched scenarios. This paper has two drawbacks: 1) It is limited in the image-based LLMs and fails to evaluate the capabilities of video LLMs; 2) GlitchBench \cite{taesiri2024glitchbench} uses \texttt{Llama-2-70b-Chat} as a judge to evaluate the model’s responses. This open-ended evaluation is unreliable and unstable due to the change in the version of the judge model. Our proposed PhysGame benchmark advances this by constructing multiple-choice questions, which facilitates more convenient evaluations. GameBugDescript \cite{taesiri2022large} is a pure-text benchmark with the reliance on pre-extracted event-wise \emph{textual} descriptions and lacks support for multi-modal evaluations. In contrast, our PhysVLM addresses all these limitations and supports multi-modal instruction evaluations in videos.
\begin{table}[t]
        \centering
        \caption{\textbf{Hyper-parameter ablations} of (a) the sampled frame number $N$ in temporal hacking and (b) the frame resolution scale factor $\gamma$ in spatial hacking for PhysDPO construction.}
    \begin{subtable}[h]{0.4\textwidth}
        \centering
        \scalebox{0.96}{
        \begin{tabular}{x{20}x{35}x{35}x{35}}
	\toprule
	\textbf{$N$} & 1 & 2 & 4   \\
	\midrule
	\textbf{AVG} & \textbf{59.5} & 58.1 & 57.8 \\
	\bottomrule
	\end{tabular}}
       \label{tab:ablateN}
    \end{subtable}
    \begin{subtable}[h]{0.4\textwidth}
        \centering
        \scalebox{0.96}{
        \begin{tabular}{x{20}x{35}x{35}x{35}}
	\toprule
        \textbf{$\gamma$} & 1/8  & 1/16 &  1/32  \\
	\midrule
	\textbf{AVG} & 57.1 & \textbf{59.5} & 58.6\\
	\bottomrule
	\end{tabular}}
        \label{tab:ablateGama}
     \end{subtable}
     \label{tab:ablaPara}
\end{table}
\begin{table*}[!ht]
\centering
\caption{\textbf{Ablations on LLMs in PhysVLM} with Vicuna-7B \cite{vicuna2023} or Qwen2-7B \cite{yang2024qwen2}.}
\vspace{-2mm}
\label{tab:ablateLLM}
\scalebox{0.9}{
\begin{tabular}{llx{22}x{20}x{20}x{20}x{20}x{20}x{20}x{20}x{20}x{20}x{20}x{20}x{20}}
\toprule
\multicolumn{1}{c}{\multirow{2}{*}{\textbf{Stage}}} & \multicolumn{1}{c}{\multirow{2}{*}{\textbf{LLMs}}} & \multicolumn{1}{c|}{\multirow{2}{*}{\textbf{AVG}}} & \multicolumn{3}{c|}{\textbf{Mechanics}} & \multicolumn{2}{c|}{\textbf{Kinematics}} & \multicolumn{3}{c}{\textbf{Optics}} & \multicolumn{4}{c}{\textbf{Material}}  \\ 
\cmidrule(r){4-6} \cmidrule(r){7-8} \cmidrule(r){9-11} \cmidrule(r){12-15}
& & & {Grav.}    & Elast.  & Fric.  & Velo.    & Acc. & Refl.   & Refr. & Abs. & Col. & Rig. & Sha. & Gest.  \\ 
\midrule
\textbf{SFT} & Vicuna & 44.7 & 47.9 & 45.0 & 48.9 & \textbf{52.1} & 48.9 & 30.4 & 42.9 & \textbf{28.8} & 28.9 & 50.0 & 31.2 & 48.3\\
\textbf{SFT} & Qwen-2 & \textbf{56.7} & \textbf{54.9} & \textbf{62.5} & \textbf{60.2} & 51.1 & \textbf{63.6} & \textbf{45.7} & \textbf{57.1}  & \textbf{28.8} & \textbf{64.4} & \textbf{51.4} & \textbf{50.0} & \textbf{72.4} \\
\midrule
\textbf{DPO} & Vicuna & 48.2 & 56.3 & 52.5 & 50.6 & \textbf{59.6} & 48.9 & 28.3 & 35.7 & 28.8 & 31.1 & 47.3 & \textbf{37.5} & 60.9   \\
\textbf{DPO} & Qwen-2 & \textbf{59.5} & \textbf{64.8} & \textbf{66.3} & \textbf{60.2} & \textbf{59.6} & \textbf{60.2} & \textbf{39.1} & \textbf{67.9}  & \textbf{35.6} & \textbf{57.8} & \textbf{62.2} & \textbf{37.5} & \textbf{78.2} \\
\bottomrule
\end{tabular}
}
\end{table*}
\begin{table*}[!ht]
\centering
\caption{\textbf{Ablations on training data} on VCG benchmark.}
\vspace{-2mm}
\scalebox{0.86}{
\begin{tabular}{llx{33}x{33}x{33}x{33}x{33}x{33}}
	\toprule
        \textbf{Stage} & \textbf{Training Data}  & \textbf{CI} & \textbf{DO} & \textbf{CU} & \textbf{TU}   & \textbf{CO} & \textbf{AVG} \\
	\midrule
        SFT & LLava-Hound & 3.48 & 2.88 & 3.74 & 2.58 & 3.02 & 3.14 \\
        SFT & LLava-Hound, LLaVA-Image & 3.43 & 2.99 & 3.73	& 2.56 & 3.12 & 3.17\\
        SFT & LLava-Hound, LLaVA-Image, PhysInstruct & \textbf{3.59} & \textbf{3.07} & \textbf{3.89} & \textbf{2.74} & \textbf{3.44} & \textbf{3.35} \\
        \midrule
        DPO & LLava-Hound-DPO  & 3.94 & 3.43 & 4.25 & \textbf{3.12} & 4.05 & 3.76\\
        DPO & LLava-Hound-DPO, PhysDPO & \textbf{3.89} & \textbf{3.69} & \textbf{4.26} & 3.11 & \textbf{4.19} & \textbf{3.83} \\
        \bottomrule
\end{tabular}}
\label{tab:ablateDataVCG}
\end{table*}
\begin{table*}[!ht]
\centering
\caption{\textbf{Ablations on training data} on Video-MME benchmark.}
\vspace{-2mm}
\label{tab:ablateDataVideoMME}
\scalebox{0.83}{
\begin{tabular}{llccccccccc} 
\toprule
\multicolumn{1}{c|}{\multirow{2}{*}{\textbf{Models}}}  & \multicolumn{1}{l}{\multirow{2}{*}{\textbf{Training Data}}} & \multicolumn{2}{c}{\textbf{Short (\%)}} & \multicolumn{2}{c}{\textbf{Medium (\%)}} & \multicolumn{2}{c}{\textbf{Long (\%)}} & \multicolumn{2}{c}{\textbf{Overall (\%)}}  \\ 
 & & \emph{w/o} subs    & \emph{w/} subs     & \emph{w/o} subs    & \emph{w/} subs    & \emph{w/o} subs   & \emph{w/} subs  & \emph{w/o} subs & \emph{w/} subs \\ 
\midrule
SFT & LLava-Hound & 65.6 & 68.9 & 55.3 & 60.4 & 47.7 & 52.4 & 56.2 & 60.6\\
SFT & LLava-Hound, LLaVA-Image & 65.2 & 68.3 & 54.9 & 60.2 & 47.6 & 52.8 & 55.9 & 60.4\\
SFT & LLava-Hound, LLaVA-Image, PhysInstruct & 64.1 & 68.0 & 55.0 &  61.7 & 46.4 & 50.3 & 55.2 & 60.0 \\
\midrule
DPO & LLava-Hound-DPO  & 66.0 & 70.2 & 53.6 & 60.5 & 47.3 & 52.8 & 55.6 & 61.2 \\
DPO & LLava-Hound-DPO, PhysDPO & 66.1 & 70.0 & 54.3 & 59.6 & 47.1 & 53.8 & 55.8 & 61.1\\
\bottomrule
\end{tabular}
}
\end{table*}


\subsection{More Ablation Studies} \label{sec:6_2}
\textbf{Ablation on LLMs in PhysVLM.} Our PhysVLM is built upon PPLLaVA \cite{liu2024ppllava} by substituting the Vicuna-7B \cite{vicuna2023} LLM with the high-performing Qwen2-7B \cite{yang2024qwen2}. To verify the necessity, we report the experimental results on PhysGame with Vicuna-7B LLM in Table \ref{tab:ablateLLM}. Experimental results demonstrate that Qwen2-7B significantly enhances the performance on physical commonsense understanding, establishing the proposed PhysVLM as a strong baseline.

\noindent \textbf{Ablations on Training Data.} We utilize the hybrid training dataset across both the SFT and DPO training stages. We perform the ablation study to assess the individual contributions of each dataset. The ablative results on PhysVLM have been presented in Table 8 of the main paper and we provide the results on Video-MME and VCG benchmarks in Table \ref{tab:ablateDataVideoMME} and Table \ref{tab:ablateDataVCG}, respectively. As shown, the introduction of PhysInstruct and PhysDPO datasets enables sustained performance improvements for PhysVLM on the PhysGame and VCG benchmarks. However, these two datasets have limited impact on the Video-MME benchmark, possibly because Video-MME places greater emphasis on long-video understanding. Despite this, given the substantial performance gains observed on the PhysGame and VCG benchmarks, we argue that the PhysInstruct and PhysDPO datasets possess substantial merit.

\noindent \textbf{Hyper-parameter Ablations} The PhysDPO dataset is constructed conditioned on misleading titles (\ie, meta information hacking), fewer frames (\ie, temporal hacking) and lower spatial resolutions (\ie, spatial hacking). For implementation, we set the sampled frame number $N$ to 1 and the frame resolution scale factor $\gamma$ to 1/16. We conduct the hyper-parameter ablation studies of $N$ and $\gamma$ in Table \ref{tab:ablaPara}. 

\subsection{Visualizations} \label{sec:6_3}
\begin{figure*}[t]
	\centering
	\begin{subfigure}[b]{\textwidth}
		\centering
		\includegraphics[width=\textwidth]{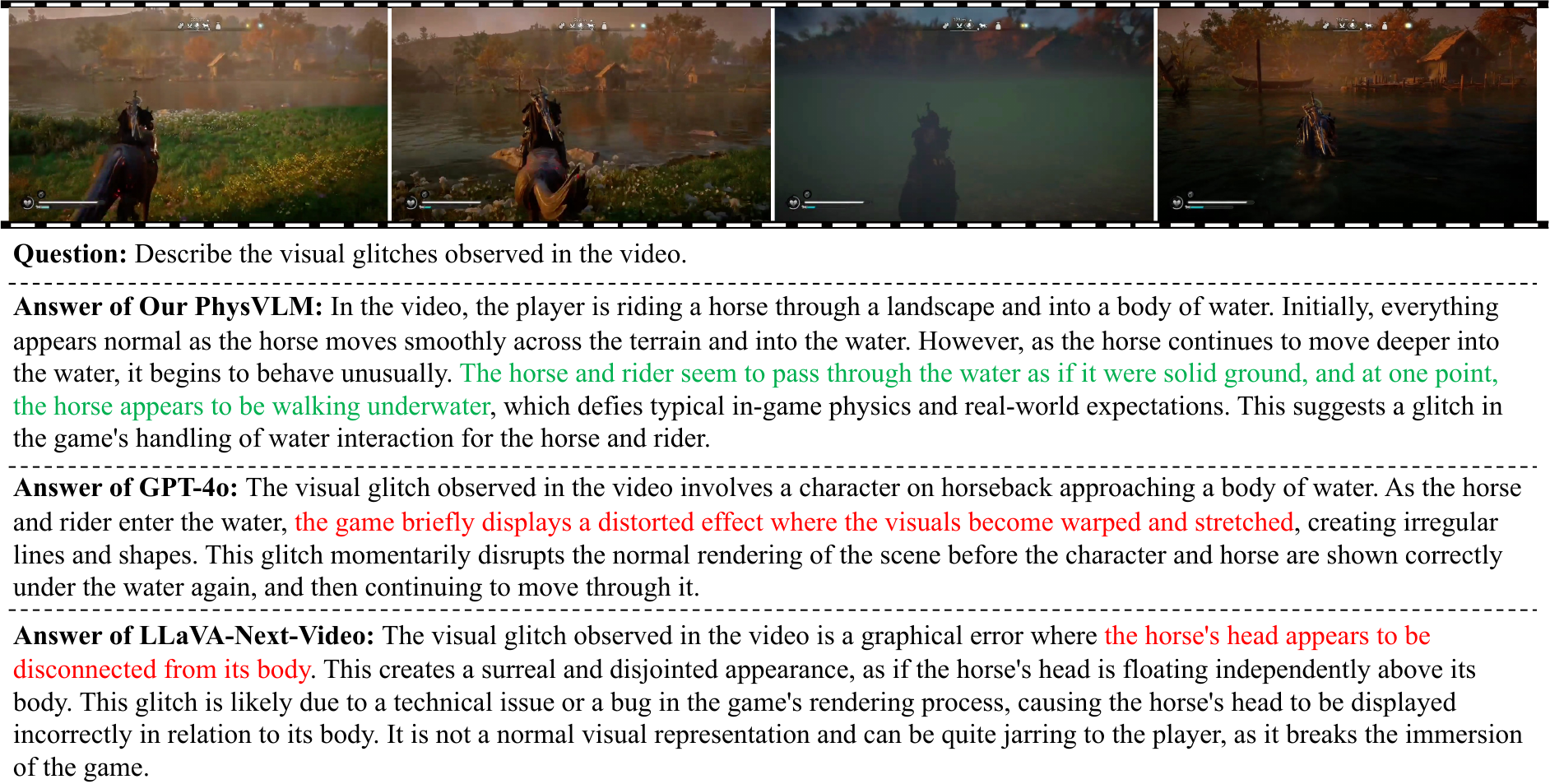}
		\caption{}
            \vspace{3mm}
	\end{subfigure}
	\begin{subfigure}[b]{\textwidth}
		\centering
		\includegraphics[width=\textwidth]{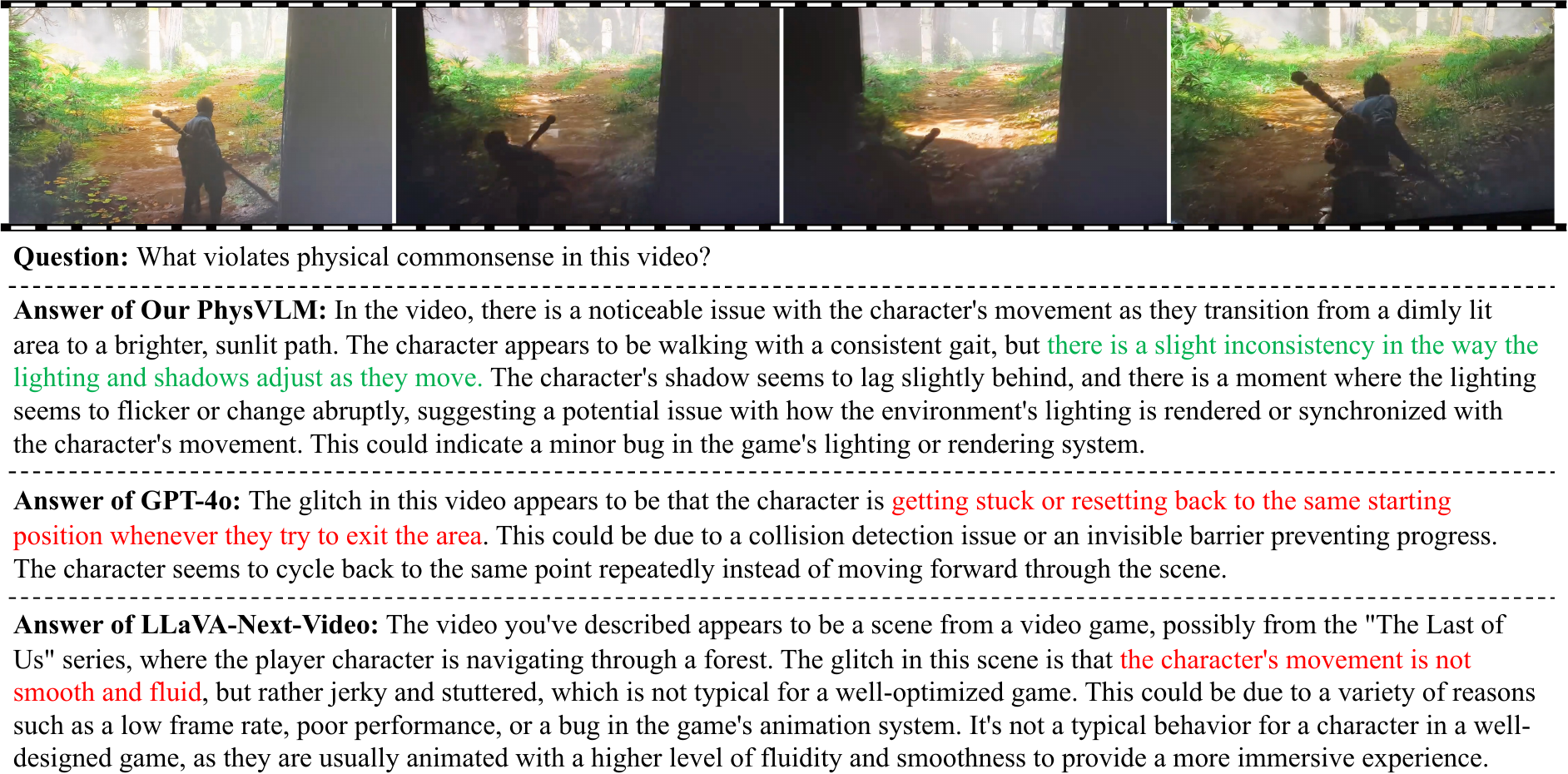}
		\caption{}
	\end{subfigure}
	\caption{\textbf{Qualitative examples} of open-ended questions.}
         \label{fig:visCompare1}
\end{figure*}
\begin{figure*}[t]
	\centering
	\begin{subfigure}[b]{\textwidth}
		\centering
		\includegraphics[width=\textwidth]{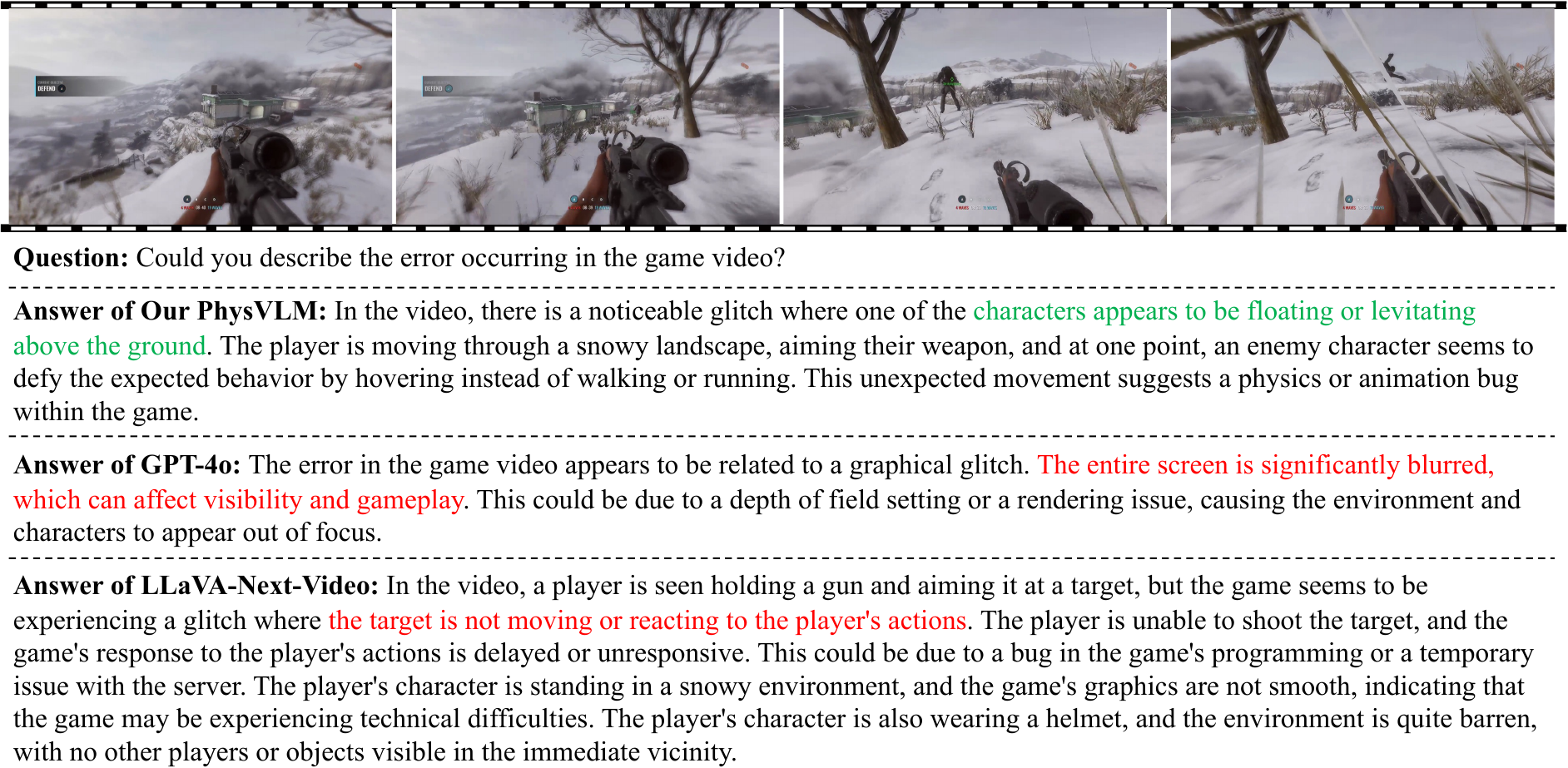}
		\caption{}
            \vspace{3mm}
	\end{subfigure}
	\begin{subfigure}[b]{\textwidth}
		\centering
		\includegraphics[width=\textwidth]{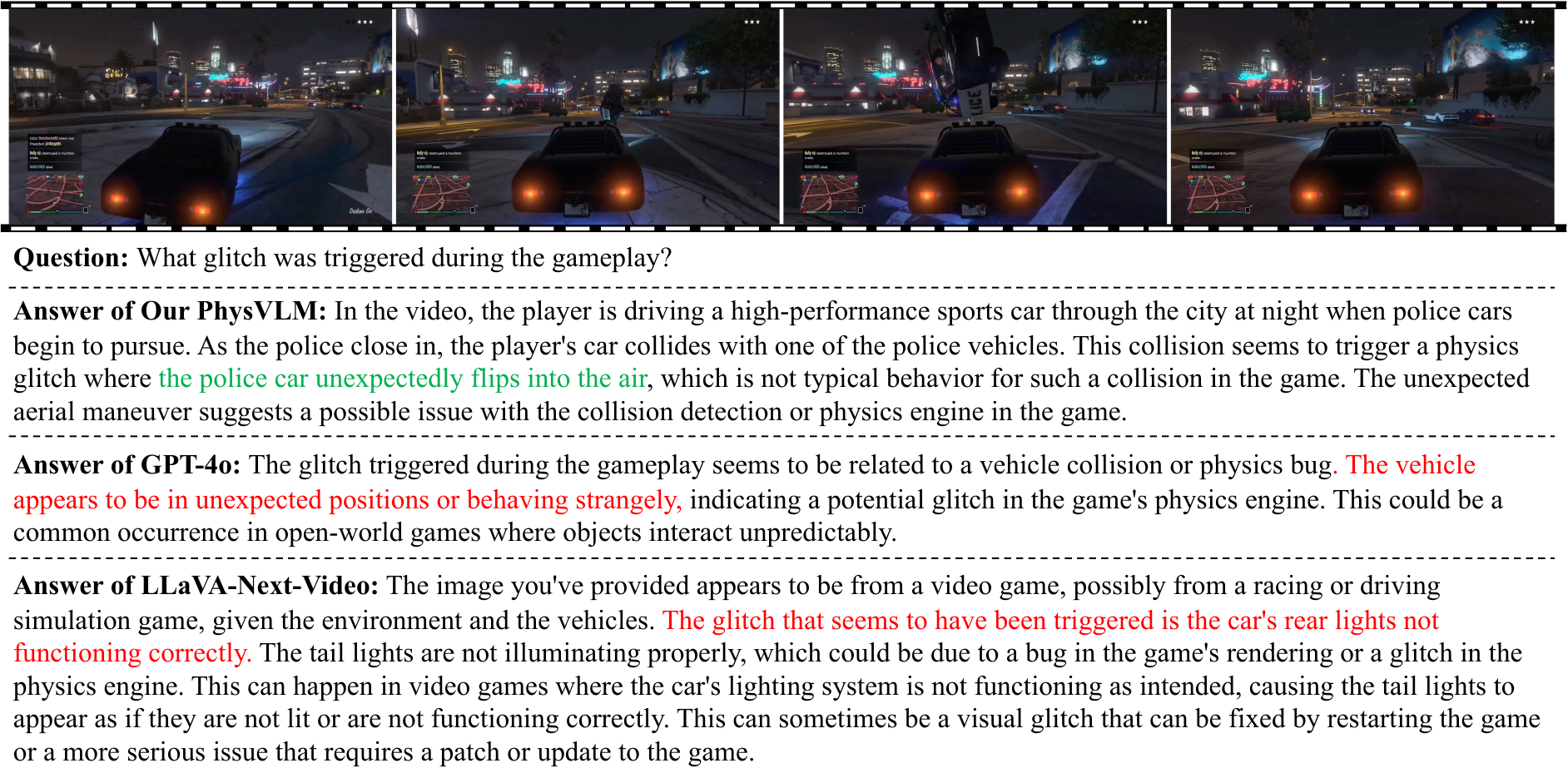}
		\caption{}
	\end{subfigure}
	\caption{\textbf{Qualitative examples} of open-ended questions.}
         \label{fig:visCompare2}
\end{figure*}
\begin{figure*}[t]
	\centering
	\begin{subfigure}[b]{\textwidth}
		\centering
		\includegraphics[width=\textwidth]{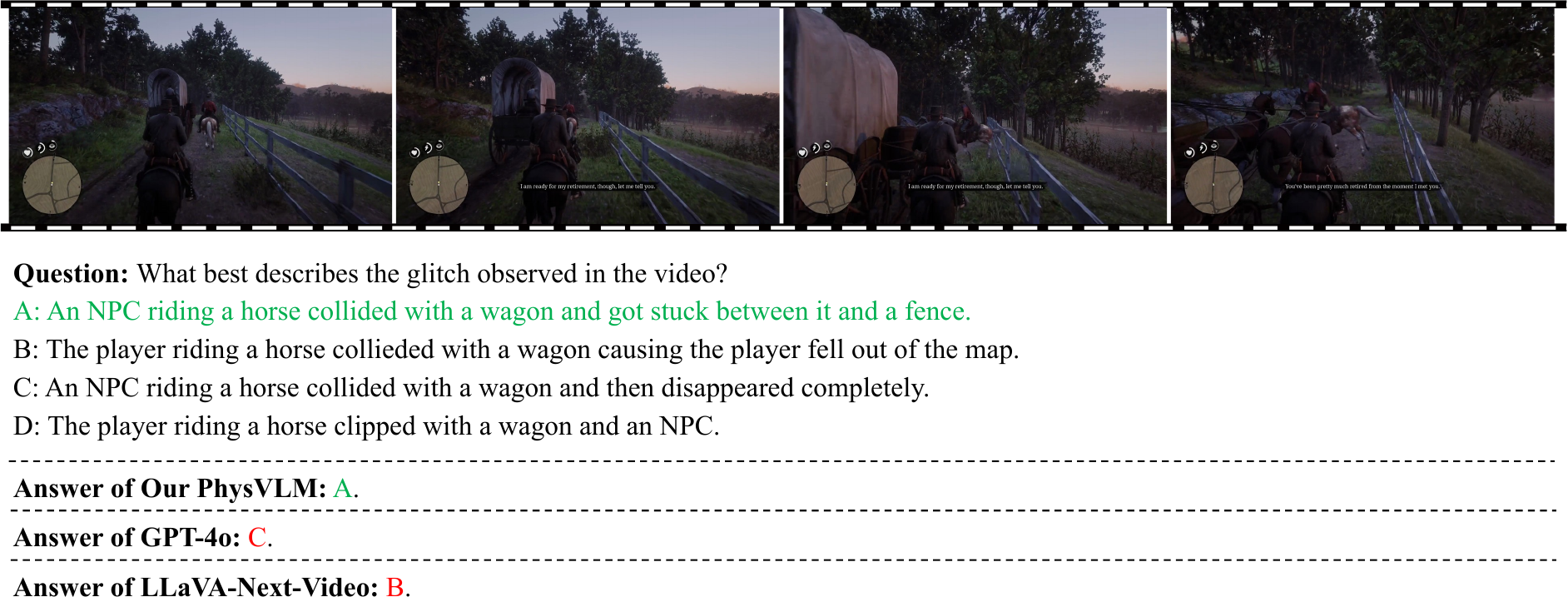}
		\caption{}
            \vspace{3mm}
	\end{subfigure}
	\begin{subfigure}[b]{\textwidth}
		\centering
		\includegraphics[width=\textwidth]{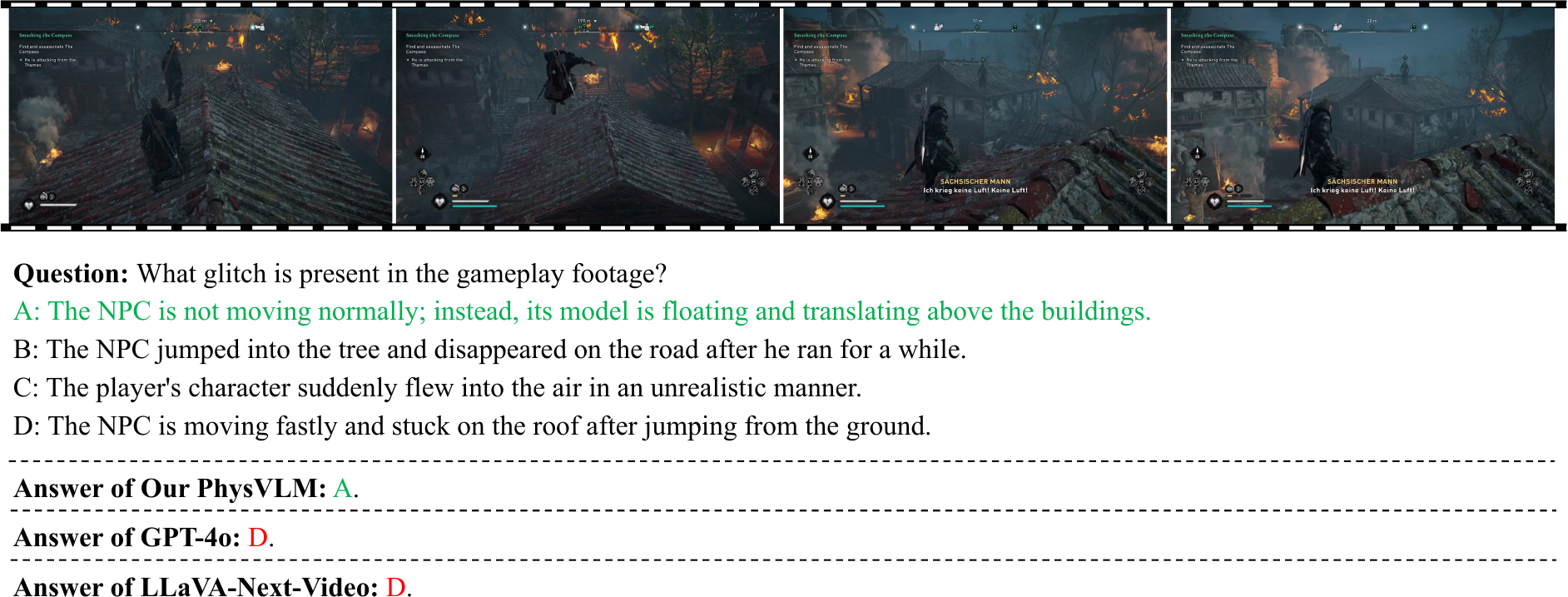}
		\caption{}
            \vspace{2mm}
	\end{subfigure}
        \begin{subfigure}[b]{\textwidth}
		\centering
		\includegraphics[width=\textwidth]{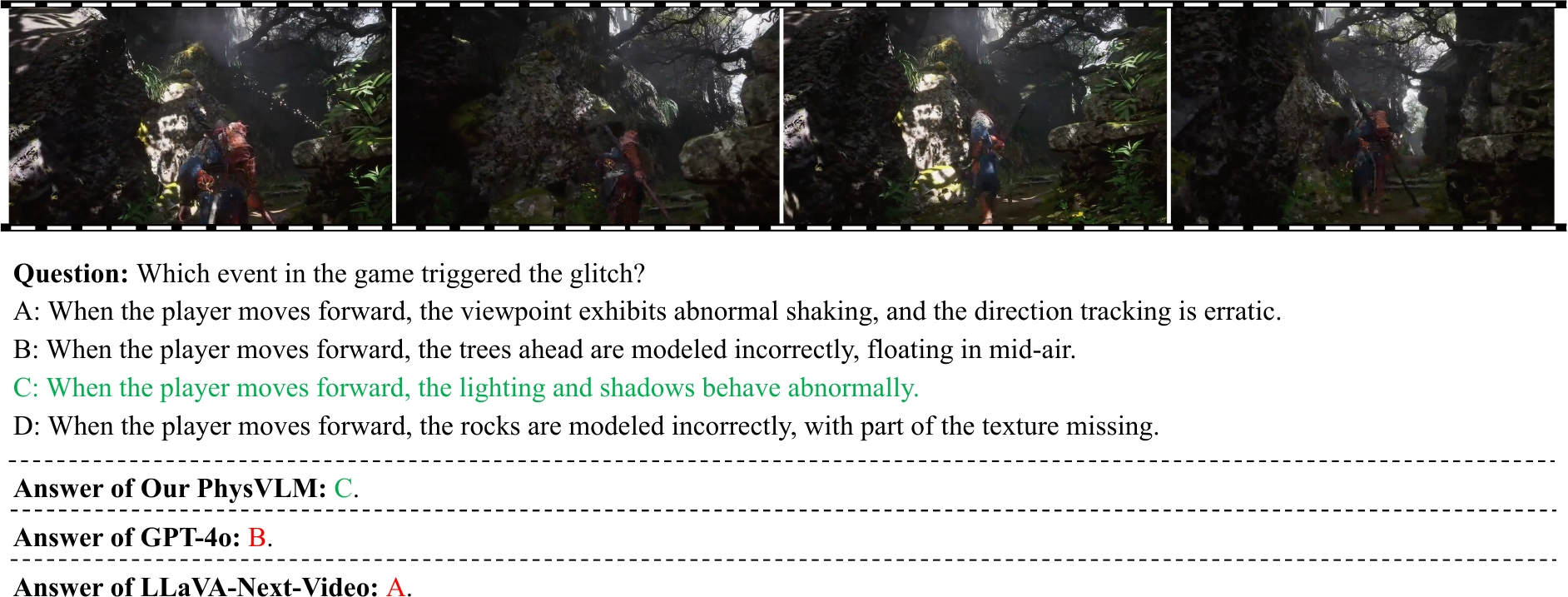}
		\caption{}
	\end{subfigure}
	\caption{\textbf{Qualitative examples} of multi-choice questions. The correct options are marked in green.}
         \label{fig:visCompareMultiChoice}
\end{figure*}
We provide the qualitative comparison results in both the formats of open-ended questions (\cf Figure \ref{fig:visCompare1} and Figure \ref{fig:visCompare2}) and multi-choice questions (\cf Figure \ref{fig:visCompareMultiChoice}). The visualization results manifest that our proposed PhysVLM effectively understands and interprets phenomena in videos that violate physical commonsense, further advancing the development of video LLMs.

\subsection{Prompts} \label{sec:6_4}
The prompts for SFT data, DPO data, and evaluation are illustrated in Table \ref{tab:sft_prompt}, Table \ref{tab:dpo_prompt}, and Table \ref{tab:evaluation_prompt}.
\begin{table*}[t]
\centering
\caption{\textbf{Prompt} for instruction-tuning data generation in PhysInstruct.}
\vspace{-2mm}
\begin{minipage}{2\columnwidth}
\vspace{0mm}    
\centering
\begin{tcolorbox}[colback=gray!5!white,colframe=black!75!black]
\centering
\begin{tabular}{p{2\columnwidth}}
\begin{minipage}{2\columnwidth}\vspace{0mm}
``role": ``system" \\
You are an AI visual assistant, and you are seeing a video and a title as a hint. Watch the video carefully and \\
analyze the events and object movements, focusing on any inconsistencies with physical laws. Please design \\
a conversation between you and the person asking about the game description and the glitch especially. \\
Example questions: \\
What is the description of the glitch observed in the video? \\
Are there any abnormalities present in the videos? \\
Please provide a description of the video content. \\

``role": ``user"\\
Title of the video: ``\{\texttt{title}\}". This hint might not be accurate. Your analysis should be based primarily on \\your own observations and understanding of the video and do not imply the title in any of your generation.\\ 
Please directly design questions like the example and answer them in detail. Ensure that all descriptions are \\ at the video level, do not refer to ``images" or ``frames".
\end{minipage}
    \end{tabular}
\end{tcolorbox}
    \label{tab:sft_prompt}
\end{minipage}
\end{table*}
\begin{table*}[t]
\centering
\caption{\textbf{Prompt} for response generation in PhysDPO. The \texttt{false}\_\texttt{title} is randomly selected from the other videos and the \texttt{question} is instantiated by the same instruction in PhysInstruct.}
\begin{minipage}{2\columnwidth}
\vspace{-2mm}    
\centering
\begin{tcolorbox}[colback=gray!5!white,colframe=black!75!black]
\centering
\begin{tabular}{p{2\columnwidth}}
\begin{minipage}{2\columnwidth}\vspace{0mm}
``role": ``system" \\
You are an AI visual assistant, and you are seeing a video and a title as a hint. Watch the video carefully and \\
analyze the events and object movements, focusing on any inconsistencies with physical laws. Please design \\
a conversation between you and the person asking about the game description and the glitch especially. \\

``role": ``user" \\
Title of the video: ``\{\texttt{false}\_\texttt{title}\}", Questions: ``\{\texttt{question}\}".  \\
Please repeat the question (Question:) and answer them in detail (Answer:). \\
Ensure that all descriptions are at the video level, do not refer to ``images" or ``frames".
\end{minipage}
    \end{tabular}
\end{tcolorbox}
\label{tab:dpo_prompt}
\end{minipage}
\end{table*}
\begin{table*}[!ht]
\centering
\caption{\textbf{Prompt} for evaluation generation in PhysGame.}
\begin{minipage}{2\columnwidth}
\vspace{-2mm}    
\centering
\begin{tcolorbox}[colback=gray!5!white,colframe=black!75!black]
\centering
\begin{tabular}{p{2\columnwidth}}
\begin{minipage}{2\columnwidth}\vspace{0mm}
Watch the video carefully and analyze the events and object movements, focusing on any inconsistencies \\
with physical laws. Identify and highlight instances where the behavior deviates from expected real-world \\
physics, and select the most accurate option to describe the detected glitch.
\\

Answer with the option letter from the given choices directly.

\end{minipage}
\end{tabular}
\end{tcolorbox}
\label{tab:evaluation_prompt}
\end{minipage}
\end{table*}


\end{document}